\documentclass[letterpaper, 10 pt, conference]{ieeeconf}  

\IEEEoverridecommandlockouts                              

\usepackage[utf8]{inputenc}
\usepackage{caption}
\usepackage{subcaption}
\usepackage{graphicx}
\usepackage{amsfonts}
\usepackage{url}

\DeclareUnicodeCharacter{1EF3}{\`y}

\title{\LARGE \bf
Planning an Efficient and Robust Base Sequence for a Mobile Manipulator Performing Multiple Pick-and-place Tasks}

\author{Jingren Xu$^{12}$, Kensuke Harada$^{12}$, Weiwei Wan$^{12}$, Toshio Ueshiba$^{2}$ and Yukiyasu Domae$^{2}$
\thanks{$^{1}$Graduate School of Engineering Science, Osaka University      {\tt\small \{ xu@hlab., harada@, wan@ \}sys.es.osaka-u.ac.jp}}%
\thanks{$^{2}$Automation Research Team, Artificial Intelligence Research Center, National Institute of Advanced Industrial Science and Technology (AIST)    {\tt\small \{ t.ueshiba, domae.yukiyasu \}@aist.go.jp}}%
}

\begin{document}

\maketitle
\thispagestyle{empty}
\pagestyle{empty}

\begin{abstract}

In this paper, we address efficiently and robustly collecting objects stored in different trays using a mobile manipulator. A resolution complete method, based on precomputed reachability database, is proposed to explore collision-free inverse kinematics (IK) solutions and then a resolution complete set of feasible base positions can be determined. This method approximates a set of representative IK solutions that are especially helpful when solving IK and checking collision are treated separately. For real world applications, we take into account the base positioning uncertainty and plan a sequence of base positions that reduce the number of necessary base movements for collecting the target objects, the base sequence is robust in that the mobile manipulator is able to complete the part-supply task even there is certain deviation from the planned base positions. Our experiments demonstrate both the efficiency compared to regular base sequence and the feasibility in real world applications.
\end{abstract}

\section{Introduction}
A mobile manipulator is the combination of a mobile base and a manipulator, which enables both locomotion and manipulation. Therefore, mobile manipulators have become of high research interest owning to their large work space and task flexibility. Especially, mobile manipulator is suitable for picking up an object and transporting it to the desired position. Fig. 1 illustrates an industrial assembly factory, currently the part-supply tasks are still highly engaged by human workers, human workers collect the required assembly parts from the containing trays and carry them to the assembly line, and then assembled by fixed base manipulators. While mobile manipulators can be applied to automate this part-supply process, owning to the ability of picking and moving. 

In order to assemble a product {\it P}, which is comprised of several types of parts {\it $P_a$}, {\it $P_b$} and {\it $P_c$}, stored in different trays {\it $tray_a$}, {\it $tray_b$} and {\it $tray_c$}, respectively, the mobile manipulator may have to move to a sequence of positions to pick up different parts from different trays. The base sequence are comprised of a series of base positions, in which the mobile manipulator is able to grasp assembly parts from the tray with avoiding self-collision and the collision between the mobile manipulator and the environment. The mobile manipulator can move to a position to pick up the parts in only one tray, or move to a position to pick up the parts stored in multiple trays, and then carries the collected assembly parts to the designated place for further assembly.

\begin{figure}
    \centering
    \includegraphics[width = \columnwidth]{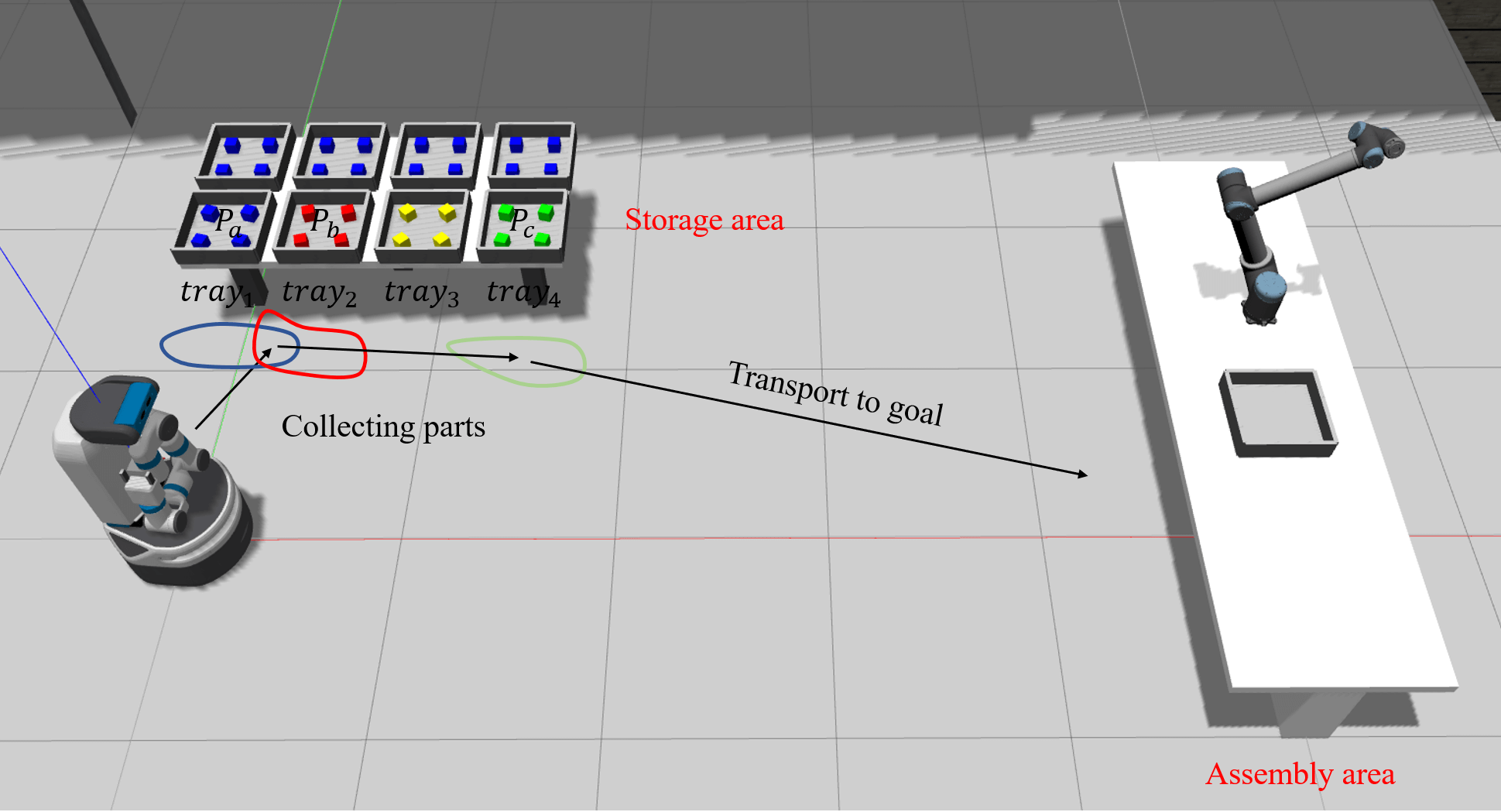}
    \caption{A schematic diagram of an assembly factory}
    \label{fig:assembly_factory}
\end{figure}

However, with increased base sequence size, the overall operation time increases significantly for the following reasons: (1) The mobile manipulator decelerates and then accelerates at every base position of the base sequence, which lowers the overall base velocity and increase the operation time. (2) Every time the mobile manipulator experiences a "stop and manipulation", there is risk that the arrived position significantly deviates from the desired position, then the mobile manipulator has to perform time-consuming repositioning process, and the risk increases with expanding base sequence. Therefore, it is important to prune out unnecessary base movements to improve the overall efficiency.

To trim the base sequence for such a part-supply task, it is preferable to move the mobile manipulator to the position, in which the mobile manipulator is able to pick up the assembly parts within multiple different trays. In Fig. 1, the mobile manipulator moves to the first base position and is able to collect the assembly parts in both $tray_1$ and $tray_2$, thus reduces the base sequence size by one. To obtain such positions, we first generate the base region, for every target tray in the given assembly task, where the mobile manipulator can grasp all the assembly parts from that tray without collision. Then the intersections of these base regions are given high priority in the base sequence planning, in addition, the base positioning uncertainty is incorporated into the base sequence planning by restricting the size of the applicable intersections, such that the planned base sequence is both efficient and robust in real world applications. 

\section{Related Work}
In this research, we aim at improving the efficiency in collecting multiple parts stored in different trays, as well as enhancing the robustness with respect to base positioning uncertainty. We presented preliminary numerical analysis on reducing the unnecessary base movements in \cite{harada2015}, while in this research, the work is conducted in a complete and feasible fashion. The completeness is achieved in by obtaining all the feasible base positions, using a resolution complete method to determine the existence of collision-free IK solutions. And the feasibility is achieved by taking into account the base positioning uncertainty of the mobile manipulator in the real environment. Moreover, the effectiveness of our proposed method is experimentally confirmed. 

Vafadar et al.\cite{vafadar2018} also researched the optimum base movements of the mobile manipulator, while they consider neither robustness nor completeness which are the main topics of this paper. There are some researches on the base placement of the mobile manipulator \cite{abolghasemi2016real}, Du et al. \cite{du2013} used the manipulability index to determine a suitable placement. OpenRAVE \cite{openrave} provides an inverse reachability module, which clusters the reachability space for a base-placement sampling distribution that can be used to find out where the robot should stand in order to perform a manipulation task. Vahrenkamp \cite{vahrenkamp2009,vahrenkamp2012,vahrenkamp2013} conducted a series of research on reachability analysis and its application, the base positions with high probability of reaching a target pose can be efficiently found from the inverse reachability distribution. The reachability indicates the probability of finding an IK solution, while there is no guarantee on the completeness of obtained base positions. Early work on the analysis of mobility and manipulability of mobile manipulator can be found in \cite{pin1990,seraji1993,yamamoto1999,tchon2000,bayle2001}, they were mainly concerned with the resolution of the kinematic redundancy caused by the mobile base, coordinated control of mobile base and the manipulator and manipulability analysis of the mobile manipulator. 

While there has been hardly any research planning the base sequence for part-supply tasks considering completeness and base positioning uncertainty. Our main contributions are:
\begin{itemize}
    \item{A resolution complete method, based on precomputed reachability database, is proposed to approximate collision free IK solutions, which is especially helpful in complex environment. And a resolution complete set of base position can be obtained by such an IK approach.}
    \item{Our reachability database does not only count the number of poses in the voxel, which gives the probability of a pose being reachable, but also stores the manipulator configurations accessible to the IK query.}
    \item{The base positioning uncertainty in real environment is considered in the base sequence planning. Following the planned base sequence, the mobile manipulator is able to efficiently and robustly collect all the parts even there is certain deviation from the planned base position.}
\end{itemize}

\section{Method Overview}

Fig. 2 shows the overview of the task. In order to assemble a specific product {\it P}, which consists of three assembly parts {\it $P_a$}, {\it $P_b$} and {\it $P_c$} stored in three different trays {\it $tray_1$}, {\it $tray_2$} and {\it $tray_4$}, we focus on planning the base sequence for collecting all the required assembly parts from the containing trays, and the following information is assumed to be known: (1) The types of parts to be collected and their associated trays. (2) The geometrical parameters of the assembly parts, trays and the potential obstacles in the environment. (3) The poses of the parts, trays and obstacles.
\begin{figure}
    \centering
    \includegraphics[width=\columnwidth]{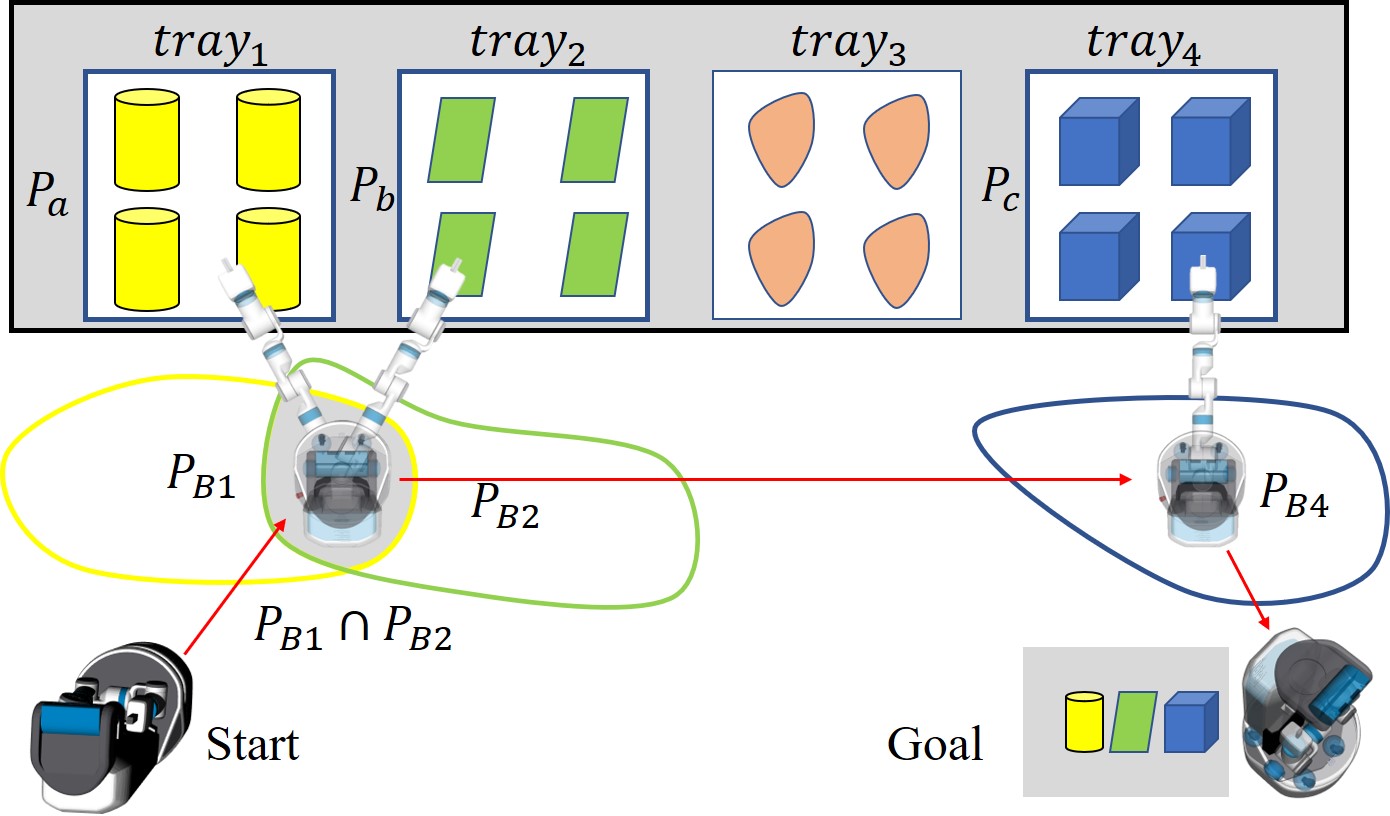}
    \caption{Overview of the method (top view). Several types of parts are required in the task, one or more parts of each type are be collected.}
    \label{fig:method_overview}
\end{figure}

We query the precomputed reachability database, for every sampled base position, if there exist collision free IK solutions that reach all the parts in one tray. The resolution complete set of base positions, where the mobile manipulator is able to grasp the parts from a tray without self-collision and the collision with environment, formulate the base region for the tray. Let $P_{Bi}$ denote the base regions of trays $tray_{i}$, $P_{Bi} \cap P_{Bj}$ denote the intersection of $P_{Bi}$ and $P_{Bj}$, the base movements can be reduced by moving the manipulator to the intersection of different base regions, for example, the mobile manipulator moves to $P_{B1} \cap P_{B2}$, and is able to pick up $P_a$ and $P_b$, then it moves to $P_{B4}$ to pick up $P_c$.

However, the intersections might be very small, the mobile manipulator will fail to collect all the parts if the actually arrived position significantly deviates from the desired base position. Considering base positioning uncertainty, we set a threshold for the size of intersections, such that the intersections are applied to reduce the number of base movements only when the radius of its inscribed circle is above the threshold, thus it is robust to a certain level of base positioning uncertainty. The threshold is determined based on base positioning performance in experiments. Then a set of robust base positions are selected from the base regions or intersections, and connected by the shortest path.

\section{Inverse Kinematics}
The inverse kinematics problem is to determine a set of joint angles that bring the end effector to a desired pose. In this study, we aim at obtaining a complete set of base positions where there exist at least one collision free IK solution that reaches desired grasping poses. Since solving IK and collision check are performed separately, this leads to a planner that loses the ability of being probabilistically complete. In terms of collision check between the mobile manipulator and the environment, it helps to find a collision-free manipulator configuration by generating a variety of candidate IK solutions that cover all kinds of manipulator configurations, thus it is not likely to miss a feasible base position in which a collision free IK solution can be found. For non-redundant manipulators, 
it is feasible to find out all the IK solutions and preform collision check between the manipulator and the environment. However, generally there are infinite number of IK solutions for redundant manipulators, a common IK solver returns only one or multiple but not necessarily representative IK solutions, in that case, the IK solver might only find IK solutions that consequently fail in collision check, even if there exist collision free solutions. Therefore, for redundant manipulators, it is important to find out representative IK solutions for further collision check. Parametrized IK approaches for 7-DOF redundant manipulator \cite{shimizu2008,singh2010} are able to find all the feasible IK solutions, while this method is usually manipulator-specific. We propose a manipulator independent method of obtaining approximated representative IK solutions, by querying the vicinity of a target pose in the reachability database, and it is proved to be a resolution complete method of solving IK.

\subsection{Reachability Database}
The reachability database is constructed by sampling the joint space of the manipulator, reachable poses are obtained by calculating forward kinematics (FK). However, this may introduce the preference of singular configurations, that large variation in joint angles only result in small difference in the grasping poses. To obtain more uniform distribution of poses in the workspace, the manipulability measure \cite{yoshikawa1985} can be applied to relieve the congestion of configurations near the singular configurations. The downsampled resultant poses of the Fetch robot \cite{fetch} are illustrated in Fig. 3(Left). The reachability can also be represented by 3 dimensional voxels containing sampled grasping poses, as shown in Fig. 3(Right), the Cartesian workspace is discretized into many 3 dimensional voxels according to the positions. The color of the voxels indicate the number of grasping poses end up into the corresponding voxels.
\begin{figure}
    \centering
    \begin{subfigure}[h]{0.23\textwidth}
        \centering
        \includegraphics[width=1\textwidth]{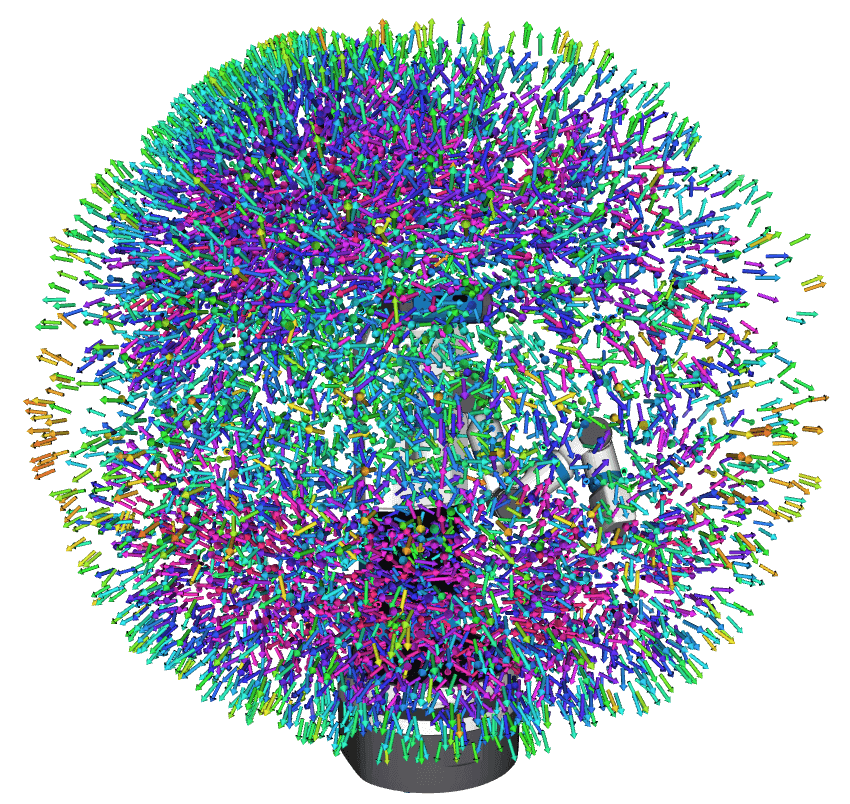}
    \end{subfigure}
    \begin{subfigure}[h]{0.23\textwidth}
        \centering
        \includegraphics[width=1\textwidth]{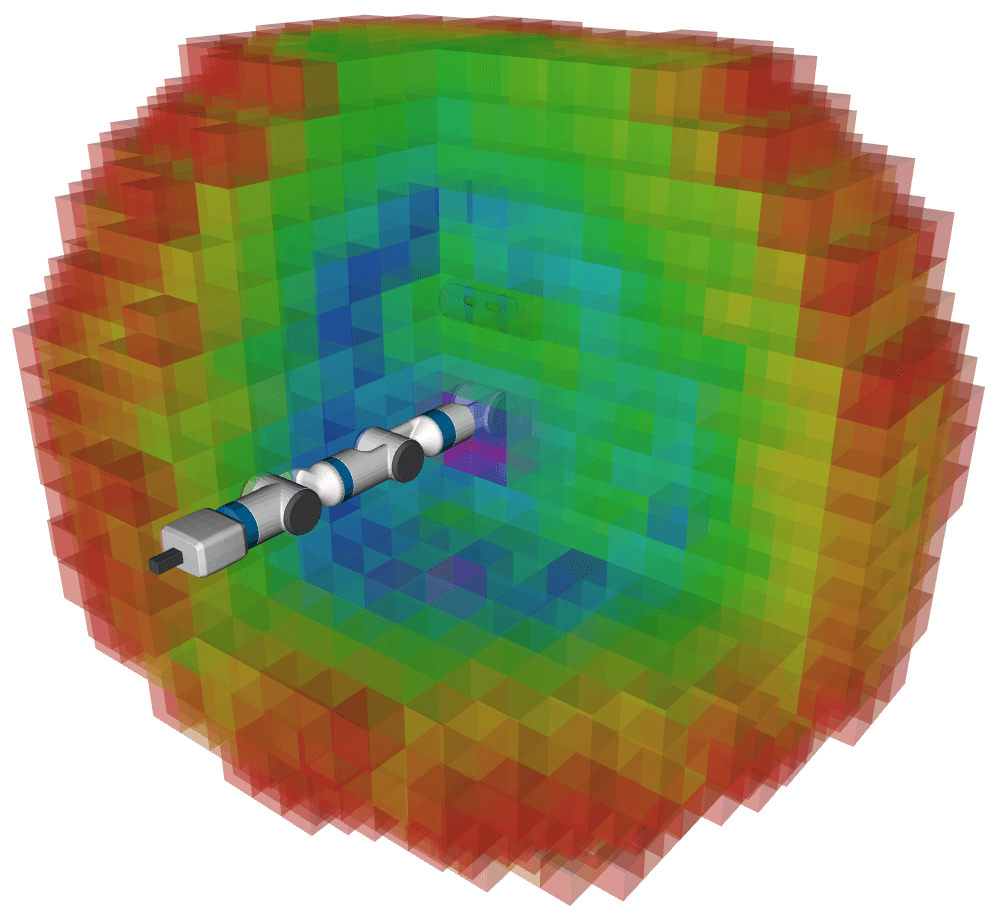}
    \end{subfigure}
    \caption{Representation of downsampled version of our reachability database. (Left) Reachable poses represented by colored arrows, the color encodes the manipulability of the corresponding manipulator configuration. (Right) The cross sectional view of the voxelized 3d workspace of a Fetch robot, the color indicates the number of grasping poses contained in the 3d voxel. For further pose query, the entire 92 million poses are distributed to 2 million 6d voxels, instead of 3d voxels.}
    \label{fig:reachability_database}
\end{figure}

For further query of poses comprised of both translation and rotation parts, 6 dimensional voxels are adopted instead, thus the workspace is discretized in both position ($x,y,z$) and orientation (represented by roll $\alpha$, pitch $\beta$ and yaw $\gamma$), the grid lengths are $\Delta x$, $\Delta y$, $\Delta z$, $\Delta \alpha$, $\Delta \beta$ and $\Delta \gamma$, respectively. All the resultant poses calculated by forward kinematics, together with their joint angles, are stored in the corresponding 6 dimensional voxels according to their positions and orientations. For example, a grasping pose $\mathcal{G}_i = [x_i, y_i, z_i, \alpha_i, \beta_i, \gamma_i]^T$, should be stored in the voxel indexed by ($x_i/\Delta x$, $y_i/\Delta y$, $z_i/\Delta z$, $\alpha_i/\Delta \alpha$, $\beta_i/\Delta \beta$, $\gamma_i/\Delta \gamma$), within the voxel is a set of poses with similar position and orientation, thus the vicinity of a target pose can be quickly found by querying the pose from such a data structure.

\subsection{IK Query}
Instead of resolving the inverse kinemtics by an IK solver, we obtain the IK solutions by querying the grasping pose in the database and access the corresponding joint angles. However, the reachability database is only a discrete representation of the continuously varying reachable poses. Probabilistically, we will fail to find an identical grasping pose in the database. Since solving IK and checking collision are treated separately, the IK solutions should be as diverse as possible in order to be resolution complete in finding collision free IK solutions. Instead, we approximate the IK solutions of the target grasping pose $\mathcal{G}_t$ by querying a range of poses in the vicinity of $\mathcal{G}_t$, for $\mathcal{G}_i \in (\mathcal{G}_t - \Delta \mathcal{G}, \mathcal{G}_t + \Delta \mathcal{G})$, where $\Delta \mathcal{G}=[\Delta x,\Delta y,\Delta z,\Delta \alpha,\Delta \beta,\Delta \gamma]^T$, and the associated manipulator configurations of $\mathcal{G}_i$ are the approximated IK solutions of $\mathcal{G}_t$.

In the reachability database, every 6d voxel contains a small range of poses, firstly we find the voxel containing the target pose, and all the poses within the voxel are regarded as the vicinity of the target pose. For example, by querying pose $[0.6,0,0.8,0,0.5,0.5]^T$ from the database, the indexed voxel is found to have 151 poses, and Fig. 4 shows a part of the manipulator configurations among them, these are the approximated IK solutions of the target pose\footnote{If exact IK solutions are desired, it is recommended to use the approximated IK solution as the initial seed in a numerical IK solver, then it can quickly converge to the exact solution after a few iterations}. As the database resolution goes to infinity, the approximation error approaches zero and the queried manipulator configurations include complete IK solutions of the target pose, such that, collision free IK solutions can be found if there exist, regardless the location of the obstacles.
\begin{figure}
    \centering
    \begin{subfigure}[h]{0.075\textwidth}
        \centering
        \includegraphics[width=1\textwidth]{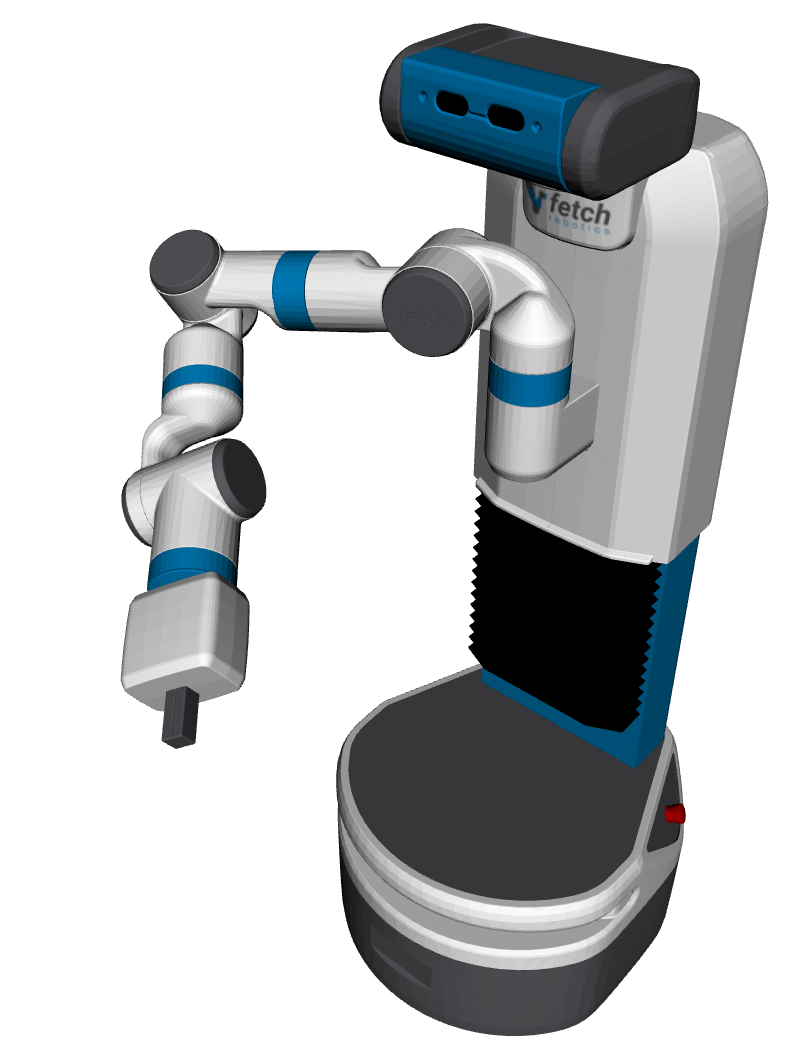}
    \end{subfigure}
    \begin{subfigure}[h]{0.075\textwidth}
        \centering
        \includegraphics[width=1\textwidth]{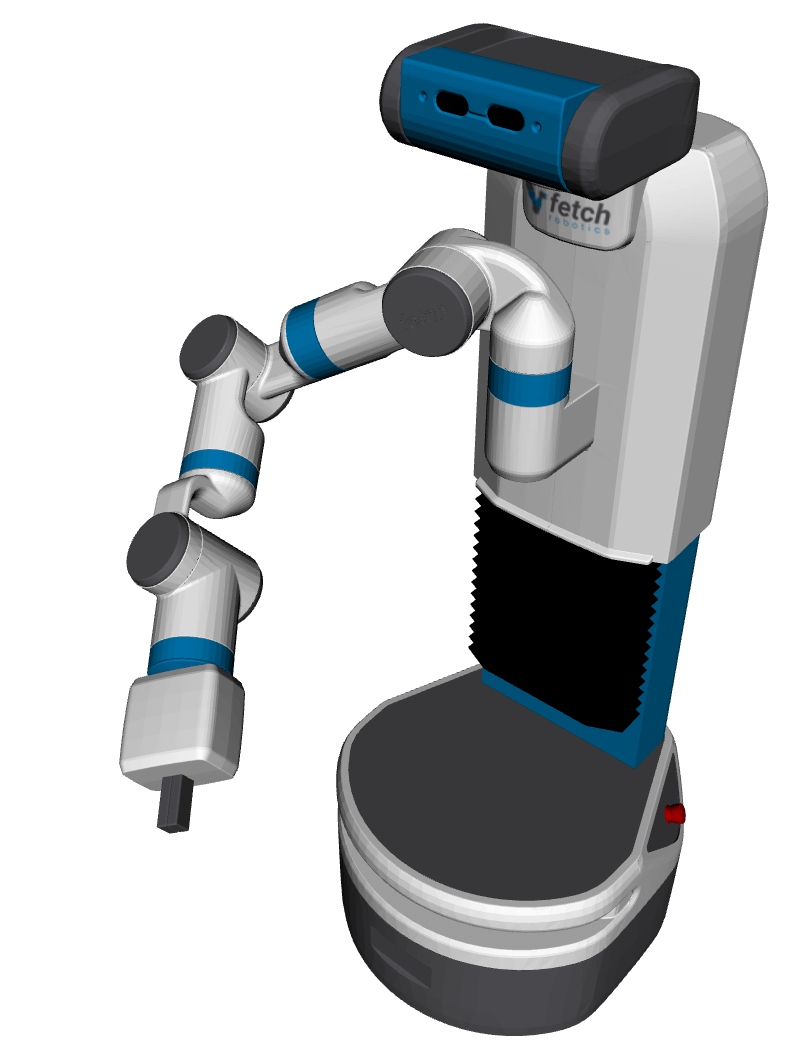}
    \end{subfigure}
    \begin{subfigure}[h]{0.075\textwidth}
        \centering
        \includegraphics[width=1\textwidth]{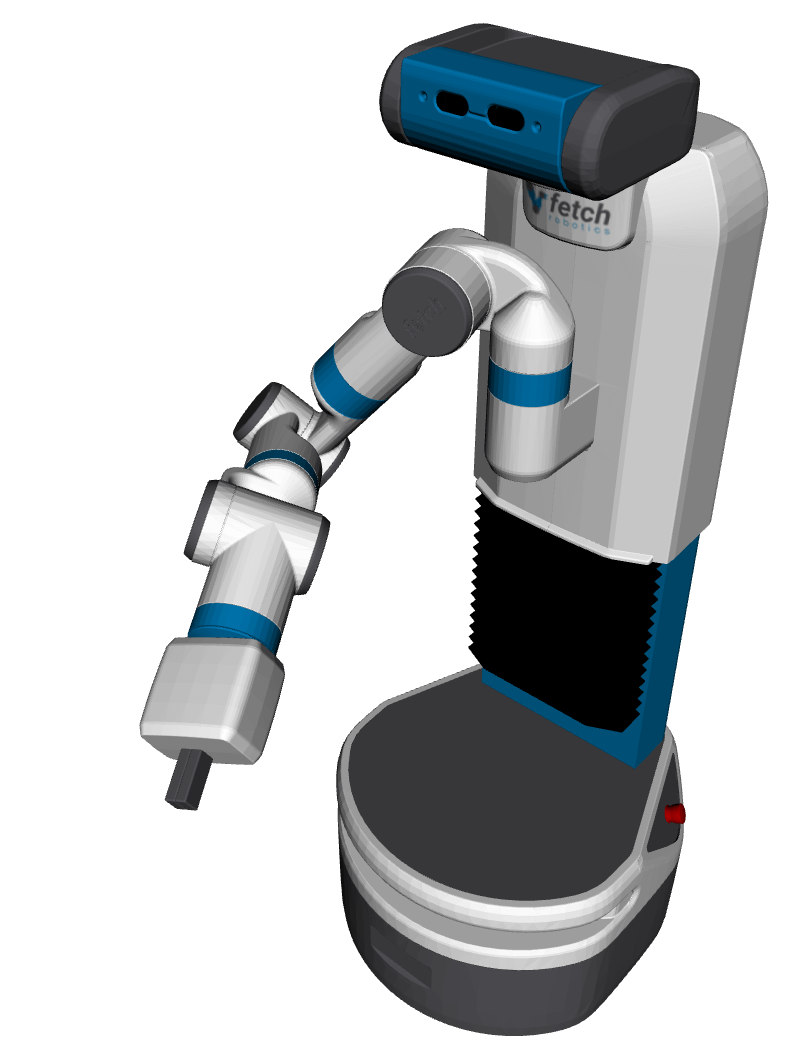}
    \end{subfigure}
    \begin{subfigure}[h]{0.075\textwidth}
        \centering
        \includegraphics[width=1\textwidth]{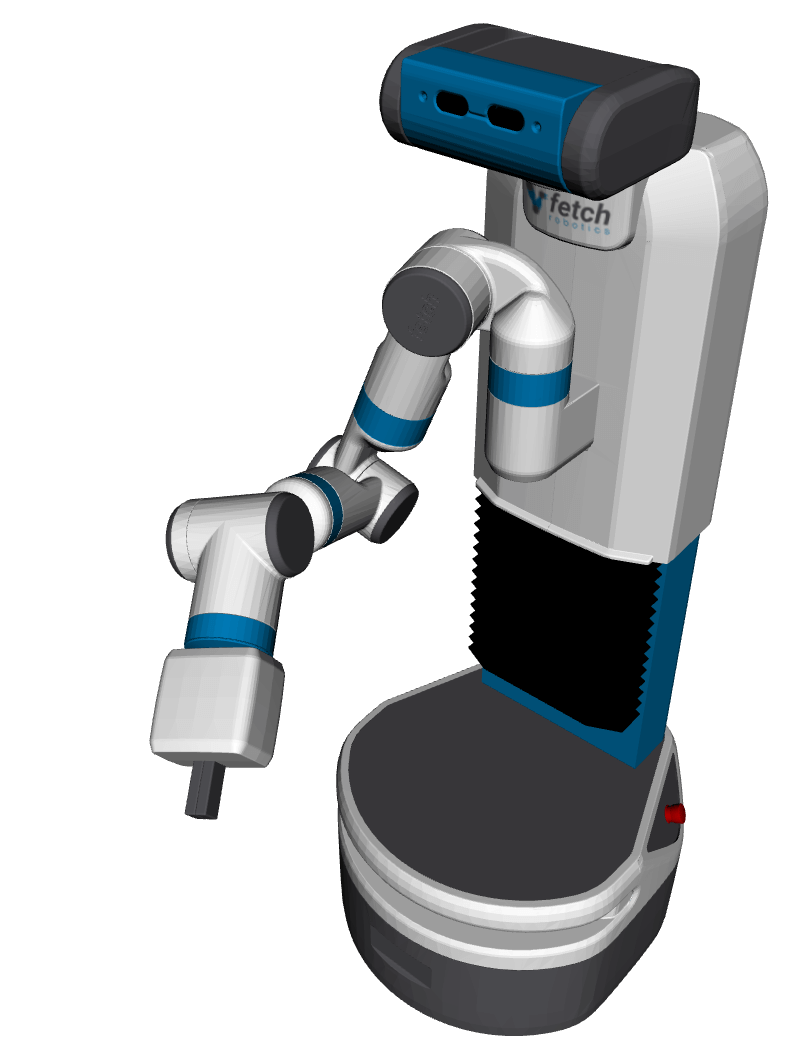}
    \end{subfigure}
    \begin{subfigure}[h]{0.075\textwidth}
        \centering
        \includegraphics[width=1\textwidth]{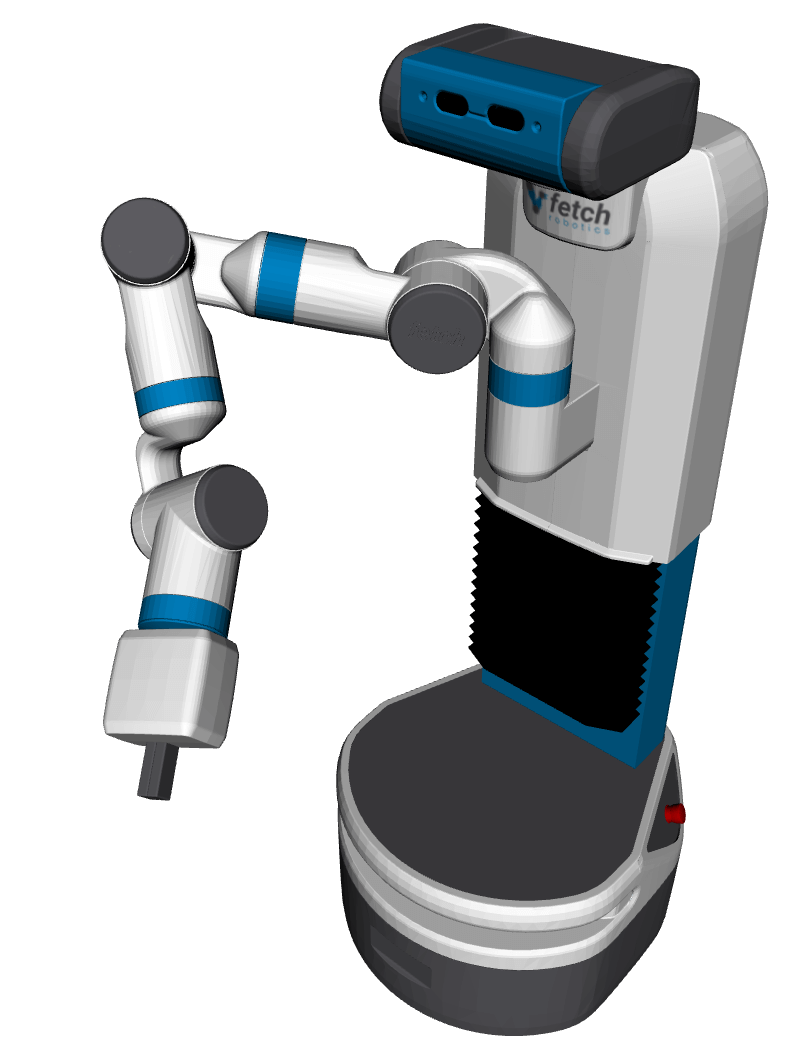}
    \end{subfigure}
    \begin{subfigure}[h]{0.075\textwidth}
        \centering
        \includegraphics[width=1\textwidth]{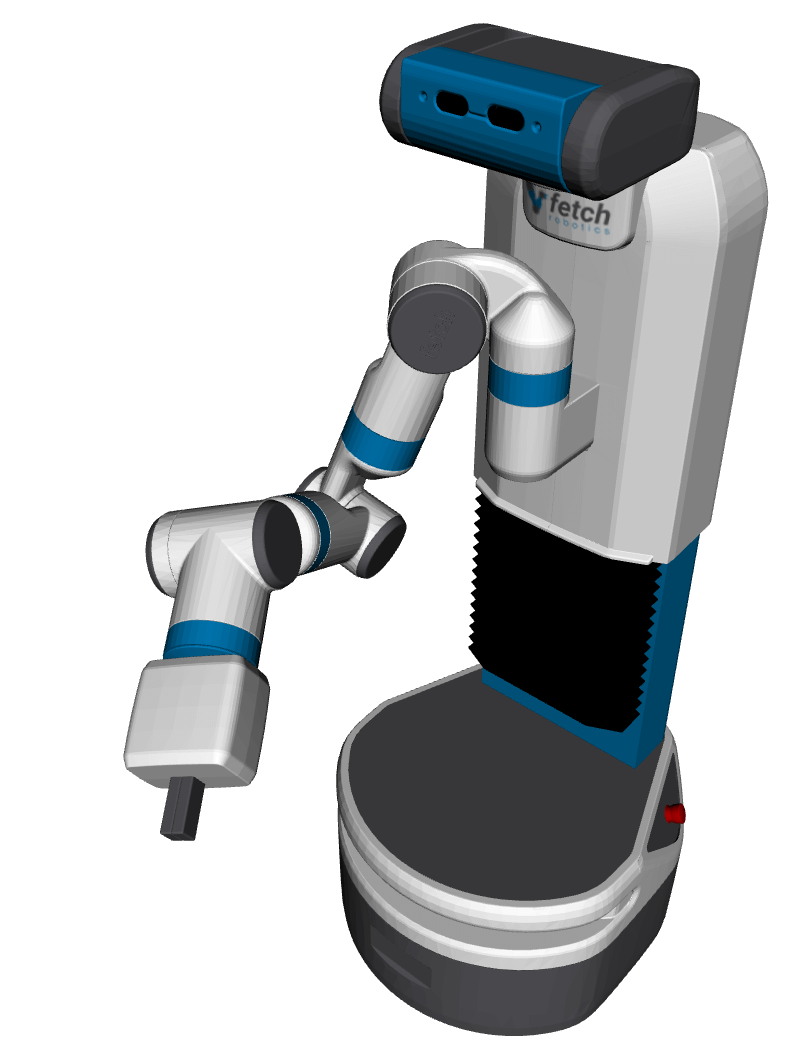}
    \end{subfigure}
    \begin{subfigure}[h]{0.075\textwidth}
        \centering
        \includegraphics[width=1\textwidth]{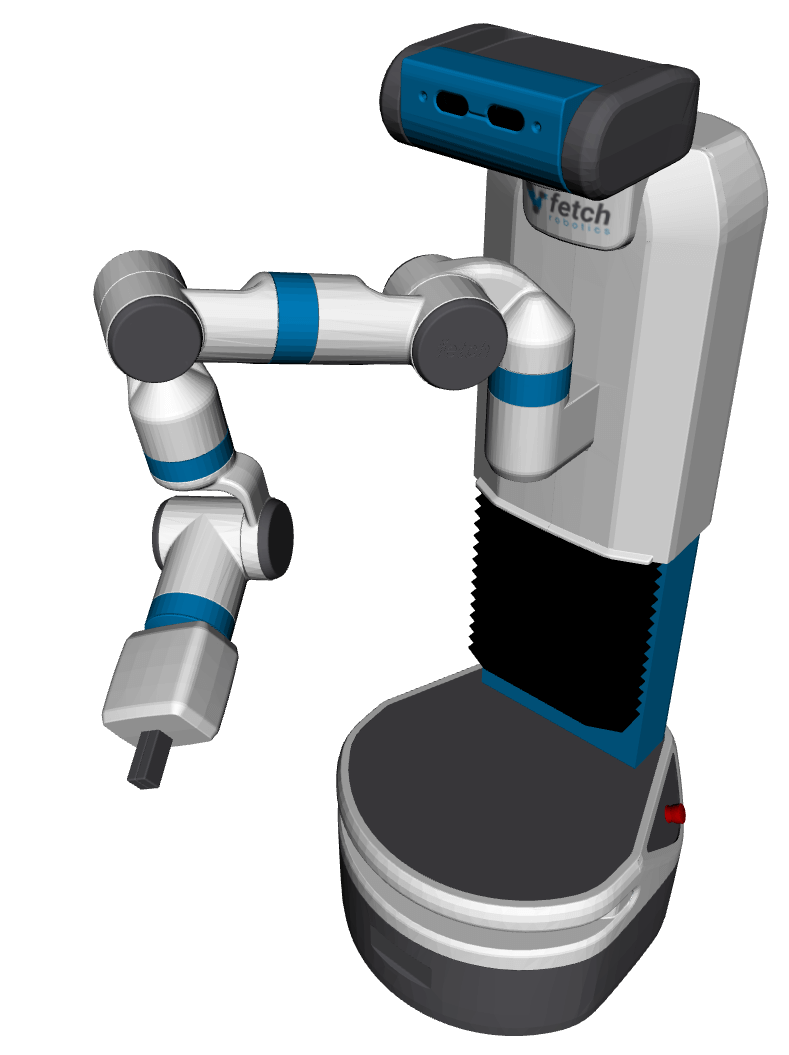}
    \end{subfigure}
    \begin{subfigure}[h]{0.075\textwidth}
        \centering
        \includegraphics[width=1\textwidth]{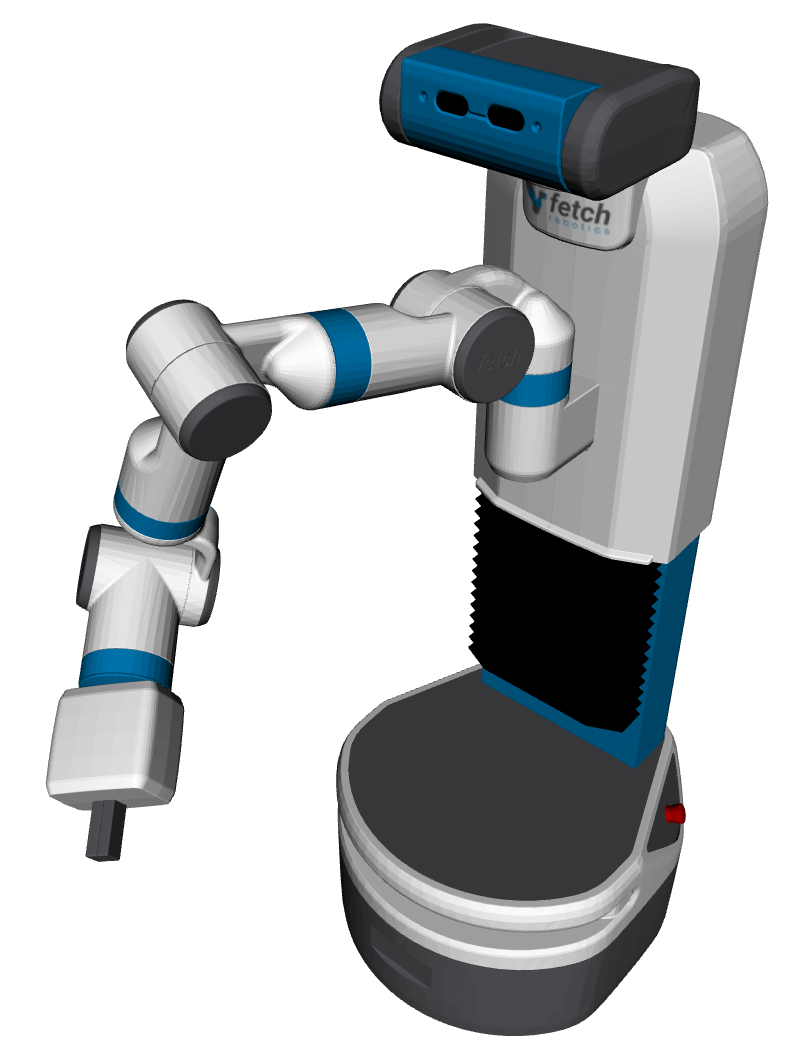}
    \end{subfigure}
    \begin{subfigure}[h]{0.075\textwidth}
        \centering
        \includegraphics[width=1\textwidth]{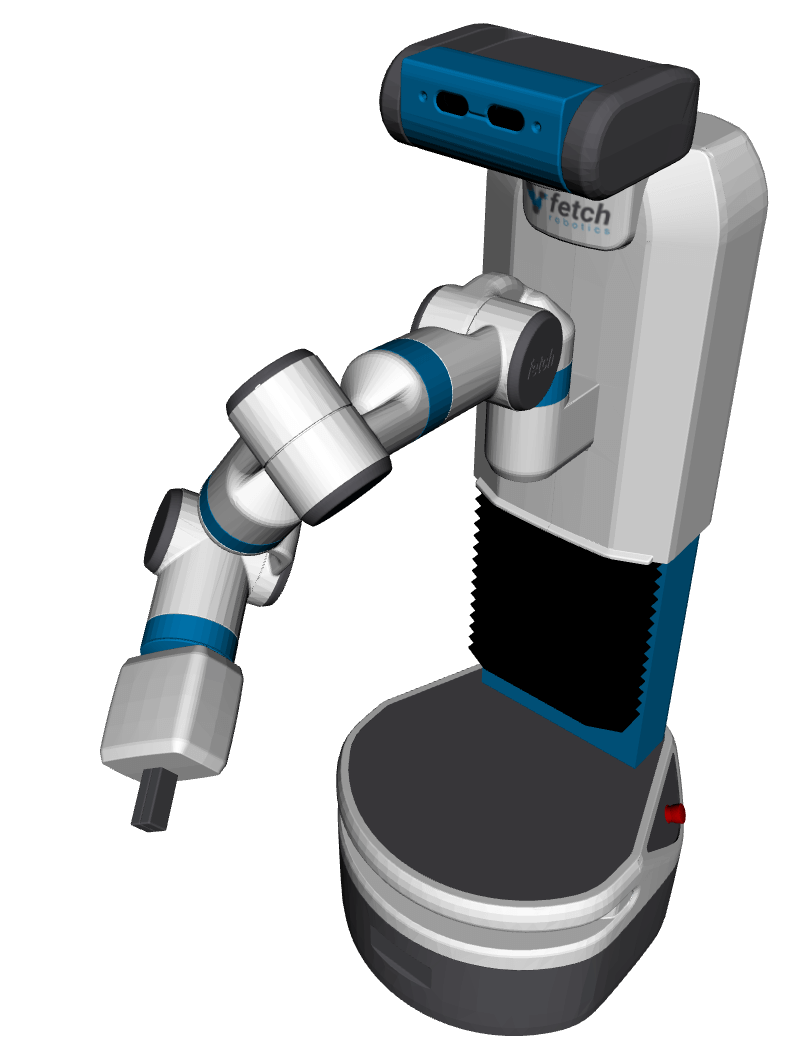}
    \end{subfigure}
    \begin{subfigure}[h]{0.075\textwidth}
        \centering
        \includegraphics[width=1\textwidth]{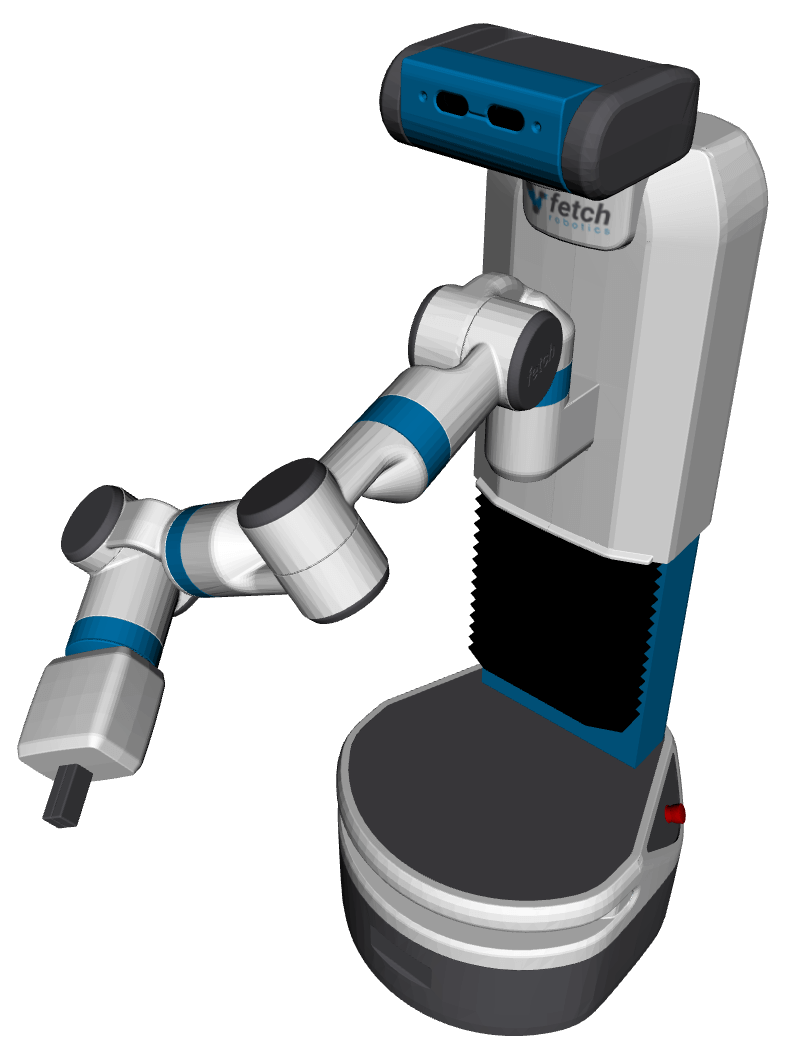}
    \end{subfigure}
    \begin{subfigure}[h]{0.075\textwidth}
        \centering
        \includegraphics[width=1\textwidth]{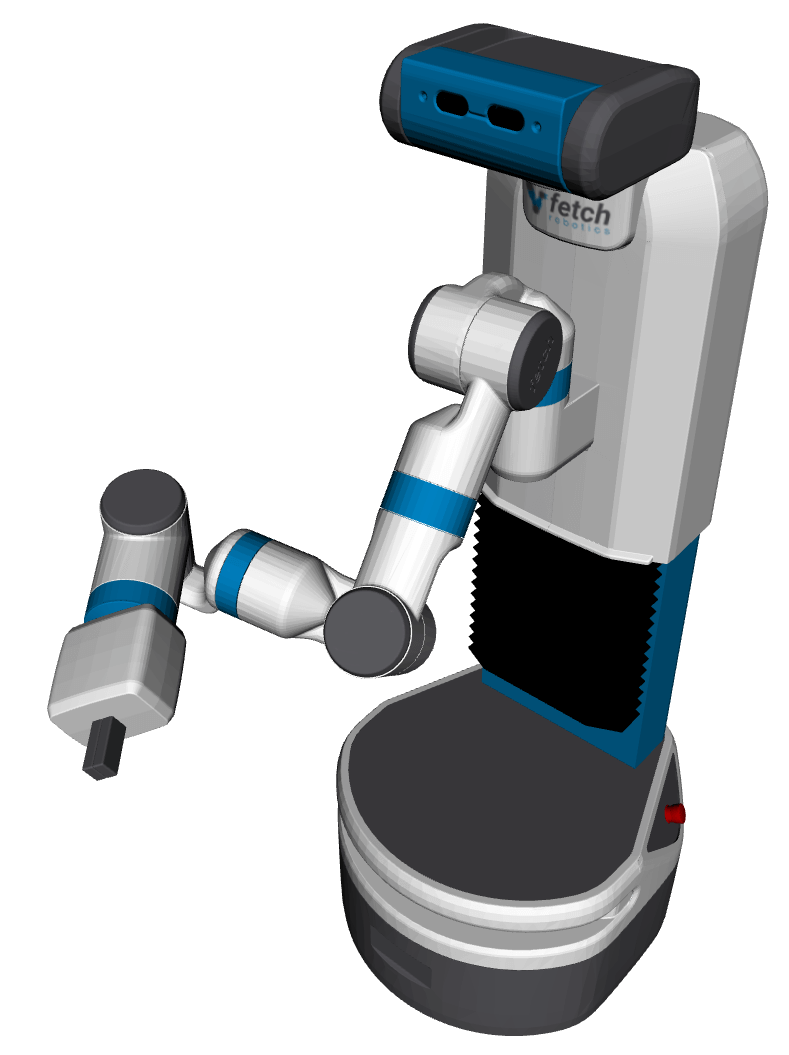}
    \end{subfigure}
    \begin{subfigure}[h]{0.075\textwidth}
        \centering
        \includegraphics[width=1\textwidth]{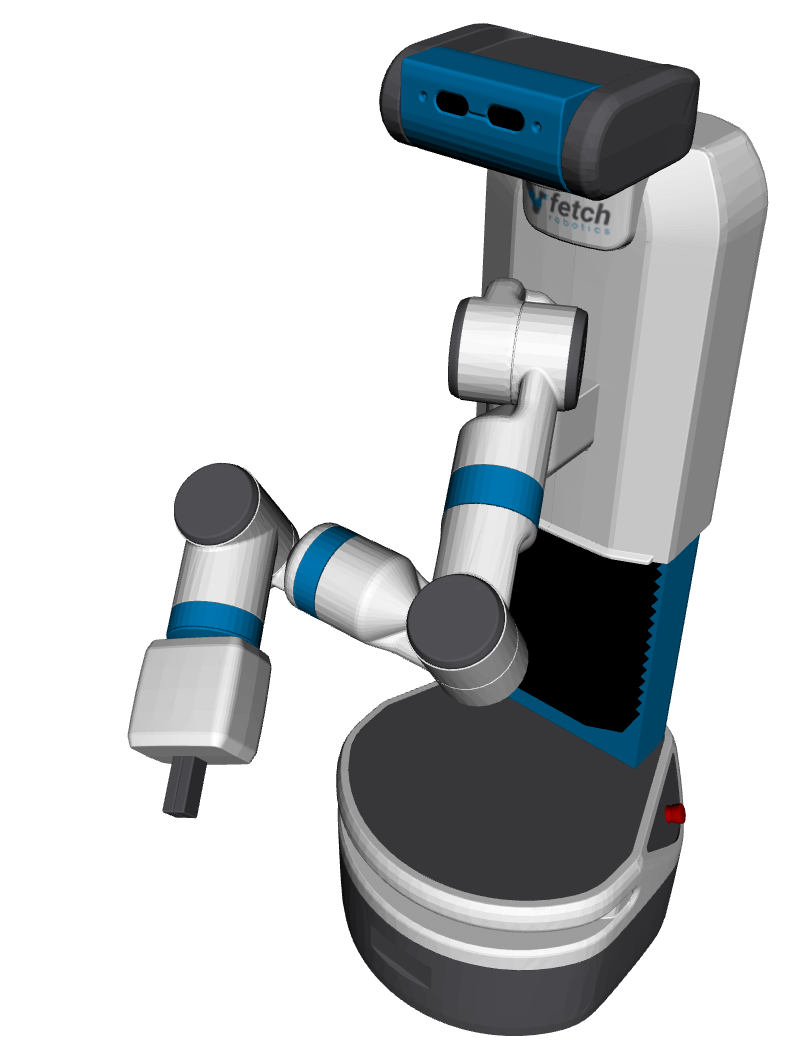}
    \end{subfigure}
    \caption{A part of obtained 151 representative manipulator configurations by querying the pose $[0.6,0,0.8,0,0.5,0.5]^T$ from the reachability database, they are the approximation of the exact IK solutions of the target pose, and cover different possible  mainipulator configurations.}
    \label{fig:representative_IKs}
\end{figure}
The feasibility and completeness of approximating IK solutions of $\mathcal{G}_t$ by the associated manipulator configurations of $\mathcal{G}_i \in (\mathcal{G}_t - \Delta \mathcal{G}, \mathcal{G}_t + \Delta \mathcal{G})$, are proved by the following two lemmas, they are based on the differentiable mapping between configuration space and workspace, except for singular configurations. The first lemma is to prove that, as the sampling resolution goes to infinity, we can always find a manipulator configuration within the voxel, that approaches any one of the IK solutions of the target pose. The second lemma proves that, as the voxelization resolution goes to infinity, all the manipulator configurations within the voxel approach the IK solutions of the target pose. Note that the completeness of approximating all the IK solutions of $\mathcal{G}_t$ is already guaranteed by lemma 1, and lemma 2, together with lemma 1, is to guarantee the completeness of obtained base positions in the next section.

\textbf{Definition:} Let $\Theta = [\theta_1,\theta_2,\dots,\theta_n]^T$ be the manipulator configuration vector, $\Delta \theta$ and $\Delta \mathcal{G}$ be the sampling and voxelization resolution, $IK(\mathcal{G}_t)$ be the IK solution set of $\mathcal{G}_t$, $FK(\Theta)$ be resultant pose calculated by forward kinematics of $\Theta$, $IKRDB(\mathcal{G}_i)$ be corresponding manipulator configuration of $\mathcal{G}_i$ in the reachability database, $J^+$ be the pseudo-inverse of Jacobian matrix $J$, $voxel(\mathcal{G}_t)$ be the voxel containing pose $\mathcal{G}_t$.

\textbf{Lemma 1:} $\forall {\Theta_t = [\theta_1,\theta_2,\dots,\theta_n]^T} \in IK(\mathcal{G}_t)$, $\exists \mathcal{G}_i \in (\mathcal{G}_t - \Delta \mathcal{G}, \mathcal{G}_t + \Delta \mathcal{G})$, $\Theta_i=IKRDB(\mathcal{G}_i)$, such that $\lim\limits_{\Delta \theta \to 0} \Vert \Theta_i - \Theta_t \Vert = 0$.

\textbf{Proof:}
Set $\{\Theta \mid FK(\Theta) = \mathcal{G}_t \}$ is equivalent to set $\{\Theta \mid \Theta = IK(\mathcal{G}_t) \}$, if $\Delta \theta \to 0$, then $\{\Theta \mid FK(\Theta) = \mathcal{G}_t\} \subset \{\Theta \mid FK(\Theta) = \mathcal{G}_i \in (\mathcal{G}_t - \Delta \mathcal{G}, \mathcal{G}_t + \Delta \mathcal{G}), \Delta \theta \to 0\}$, thus $\forall{\Theta_t} \in IK(\mathcal{G}_t)$, $\exists \mathcal{G}_i \in (\mathcal{G}_t - \Delta \mathcal{G}, \mathcal{G}_t + \Delta \mathcal{G})$, $\Theta_i=IKRDB(\mathcal{G}_i)$, such that 
$\lim\limits_{\Delta \theta \to 0} \Vert \Theta_i - \Theta_t \Vert = 0$.

\textbf{Lemma 2:} $\forall{\mathcal{G}_i} \in (\mathcal{G}_t - \Delta \mathcal{G}, \mathcal{G}_t + \Delta \mathcal{G})$, $\Theta_i=IKRDB(\mathcal{G}_i)$, $\exists \Theta_t \in IK(\mathcal{G}_t)$, that $\lim\limits_{\Delta \mathcal{G} \to 0} \Vert \Theta_i - \Theta_t\Vert  = 0$.

\textbf{Proof:} $\dot{\mathcal{G}} = J(\Theta)\dot{\Theta}$, $\dot{\Theta} = J^{+}(\Theta)\dot{\mathcal{G}}+(I-J^{+}J)k $, where $k$ is an arbitrary vector denoting redundancy, integrate two sides of the formula by a small time step, $\Delta \Theta = J^{+}(\Theta)\Delta \mathcal{G}+(I-J^{+}J)k\Delta t$, then $\forall{\Delta \mathcal{G}}\to 0$, $\exists k =0$ such that $\Delta \Theta \to 0$. Because $\vert \mathcal{G}_i - \mathcal{G}_t \vert < \Delta \mathcal{G}$, $\Delta \mathcal{G} \to 0 \Rightarrow \vert \mathcal{G}_i - \mathcal{G}_t \vert \to 0$, thus $\lim\limits_{\vert \mathcal{G}_i - \mathcal{G}_t \vert \to 0} \Vert \Theta_i - \Theta_t\Vert  = 0$. (replace $J^+$ with $J^{-1}$ for non-redundant manipulators)

The above lemmas apply to both redundant and non-redundant manipulators, the only problem with this method is that the continuous mapping between joint space and workspace breaks down at singular configurations.


\section{Base Sequence Planning}
\subsection{Base Region Calculation}
The base region for a tray is a set of base positions where the mobile manipulator is able to reach all the targets in the tray, with avoiding self-collision and the collision with the environment. Firstly, we prepare stable grasping poses with respect to the object for every object in the tray, using a grasp planner \cite{harada2008,uto2013,tsuji2014,harada2014}. Then sample base poses $(x_i, y_i, \phi)$ in front of the target tray, as illustrated in Fig. 5, here we assume that the orientation $\phi$ of the mobile manipulator is constant, because for many mobile manipulators, the joint connecting the manipulator and mobile base rotates around a vertical axis, thus counteracts the rotation of the mobile base and contributes almost nothing new. Then the set of stable grasping poses $\{\mathcal{G}_{t1},\mathcal{G}_{t2},\dots,\mathcal{G}_{tn}\}_j$ with respect to the mobile base for object $\mathcal{O}_j$ is obtained for every object in the tray. Finally, if there exists an grasping pose $\mathcal{G}_{ti} \in \{\mathcal{G}_{t1},\mathcal{G}_{t2},\dots,\mathcal{G}_{tn}\}_j$, such that we can find a collision free manipulator configuration $IKRDB(\mathcal{G}_k)$ for a pose $\mathcal{G}_k \in voxel(\mathcal{G}_{ti})$, then $\mathcal{O}_j$ can be grasped from the base position, a base position for the tray is feasible if all the objects in the tray can be grasped. All of the feasible positions formulate the base region for grasping objects from the tray.

We compare the base regions obtained by different IK approaches. The obtained base regions using IKFast plugin to find the IK solutions are shown in Fig. 6. Fig. 7 is the base regions calculated by IK query approach proposed in this paper, it is obvious the base region is larger. In the IK solver approach, IK solutions are not found in some base positions, and some of the found IK solutions fail in the collision check. For the IK query approach, the obtained base positions are not always feasible, but the base region is complete when the resolution of reachability database goes to infinity.
\begin{figure}
    \centering
    \includegraphics[width=0.4\textwidth]{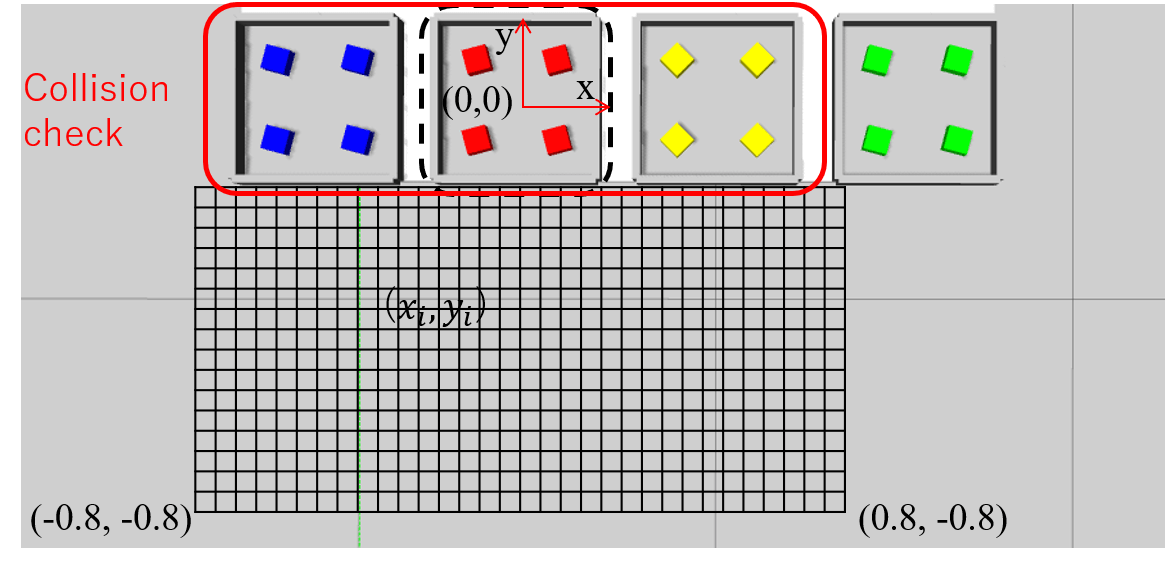}
    \caption{The sampled base positions for one of the trays, circled by black dashed line. The range of sampling is determined referring to reachable workspace in Fig. \ref{fig:reachability_database}. Collision check is performed between the mobile manipulator and the target tray together with its neighboring trays.}
    \label{fig:base_sampling}
\end{figure}
\begin{figure}
    \centering
    \begin{subfigure}[h]{0.23\textwidth}
        \centering
        \includegraphics[width=1\textwidth]{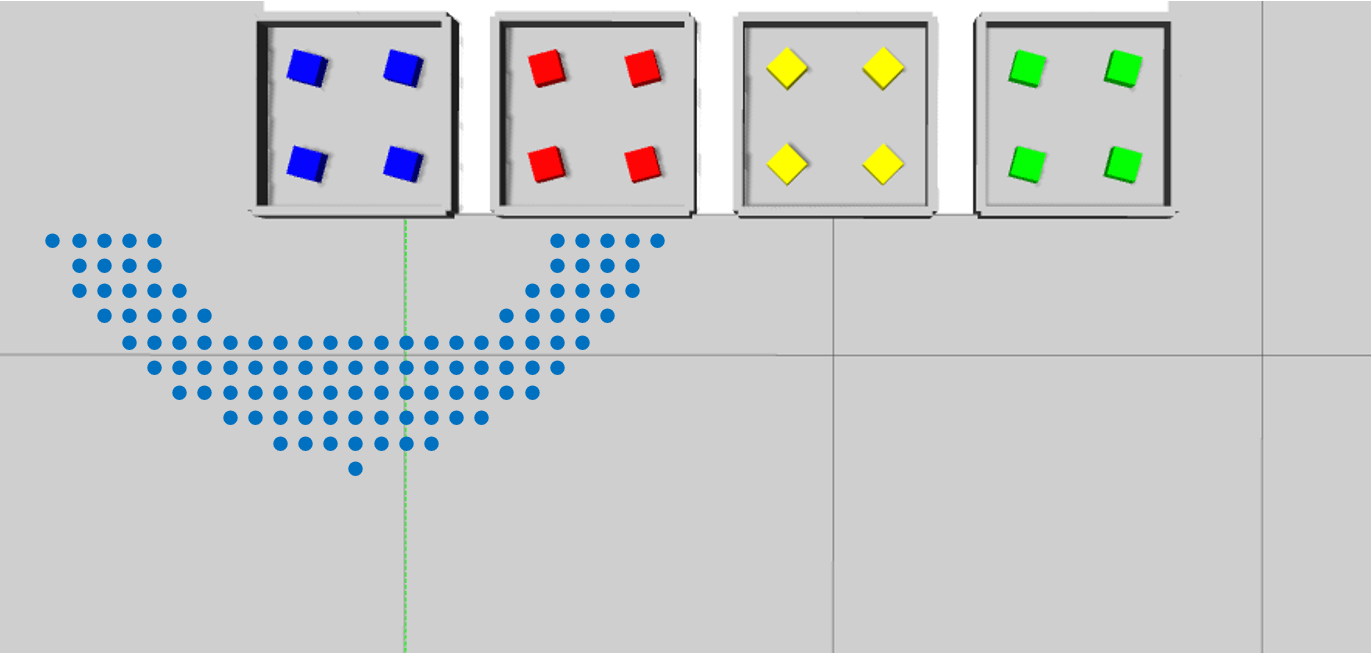}
    \end{subfigure}
    \begin{subfigure}[h]{0.23\textwidth}
        \centering
        \includegraphics[width=1\textwidth]{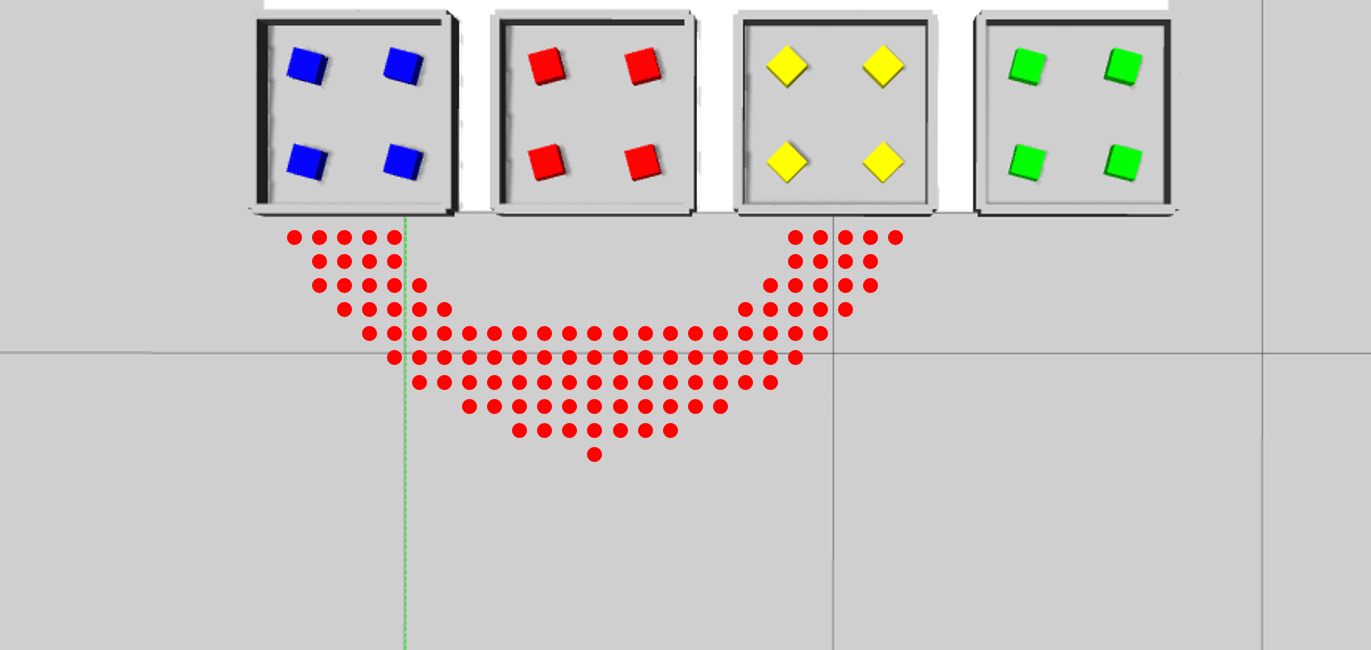}
    \end{subfigure}
    \begin{subfigure}[h]{0.23\textwidth}
        \centering
        \includegraphics[width=1\textwidth]{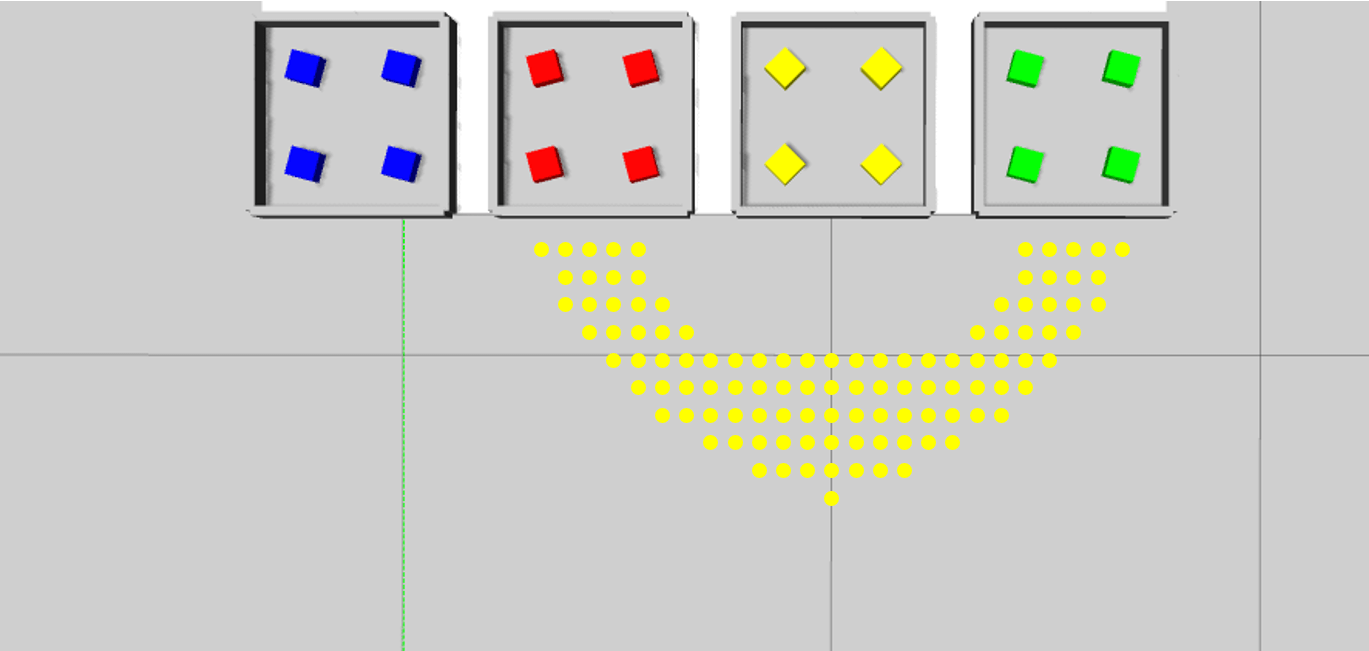}
    \end{subfigure}
    \begin{subfigure}[h]{0.23\textwidth}
        \centering
        \includegraphics[width=1\textwidth]{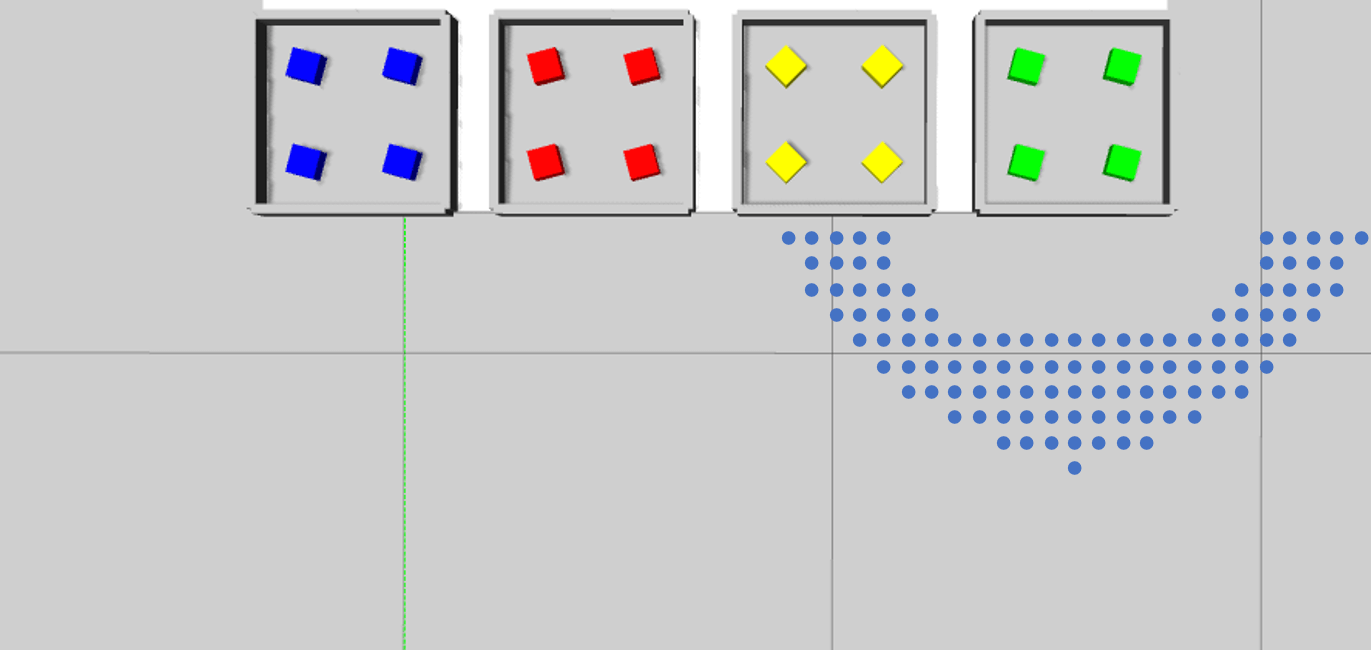}
    \end{subfigure}
    \caption{The base regions obtained by using the IKFast solver.}
    \label{fig:base_regions_IKsolver}
\end{figure}
\begin{figure}
    \centering
    \begin{subfigure}[h]{0.23\textwidth}
        \centering
        \includegraphics[width=1\textwidth]{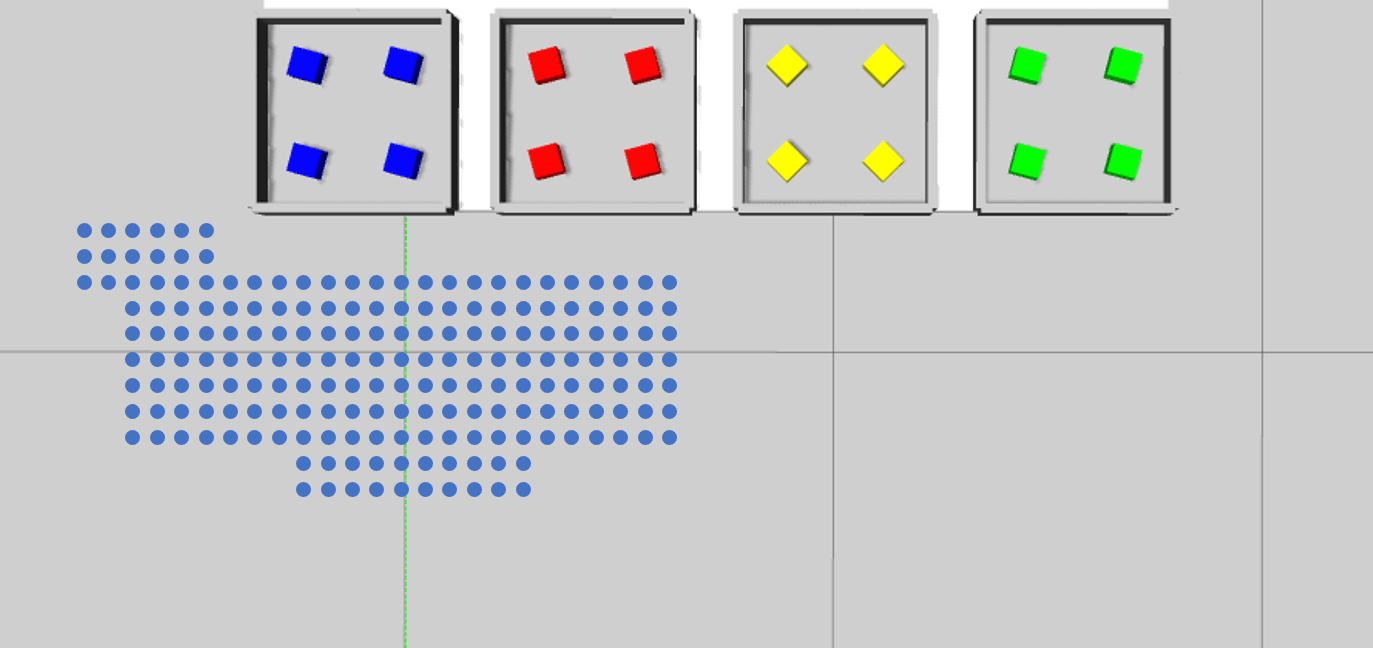}
    \end{subfigure}
    \begin{subfigure}[h]{0.23\textwidth}
        \centering
        \includegraphics[width=1\textwidth]{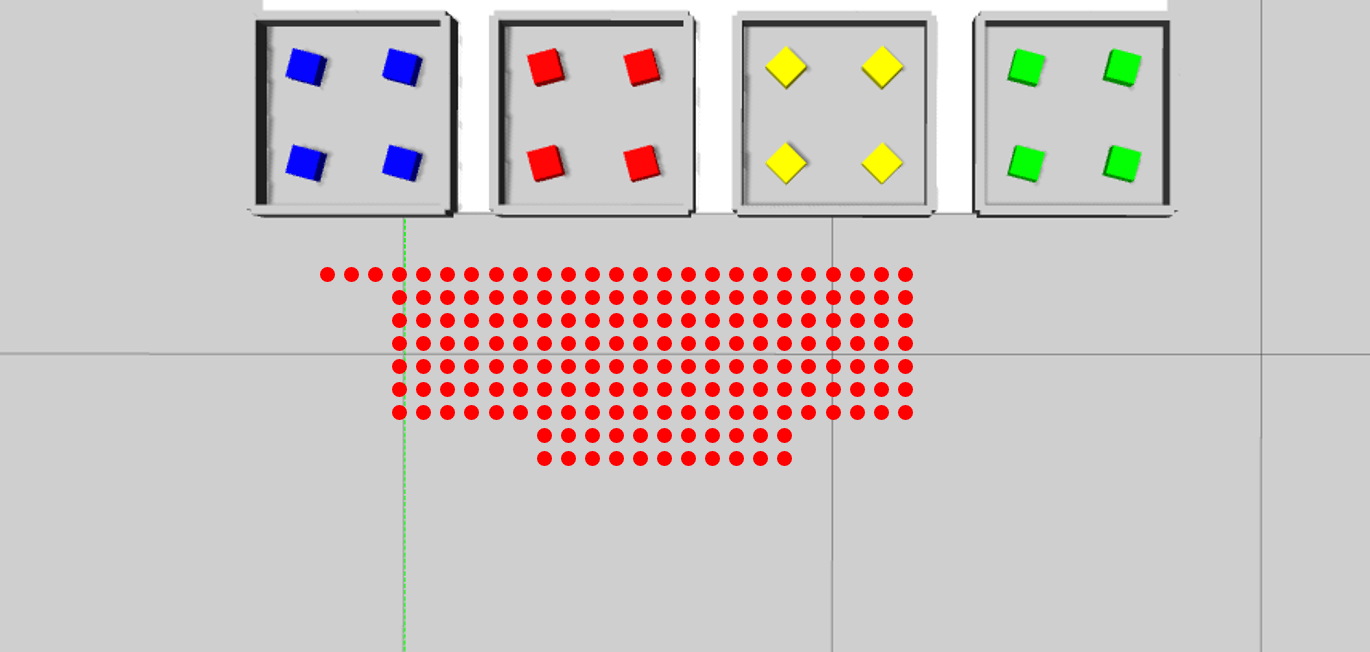}
    \end{subfigure}
    \begin{subfigure}[h]{0.23\textwidth}
        \centering
        \includegraphics[width=1\textwidth]{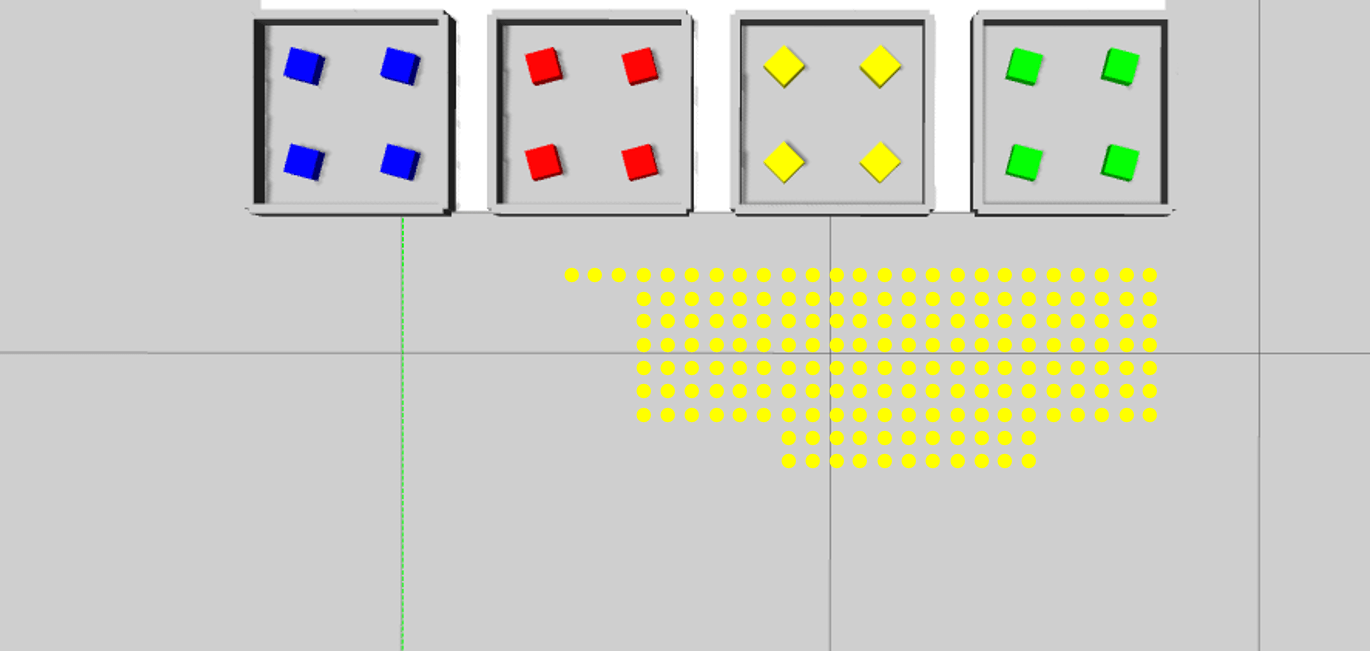}
    \end{subfigure}
    \begin{subfigure}[h]{0.23\textwidth}
        \centering
        \includegraphics[width=1\textwidth]{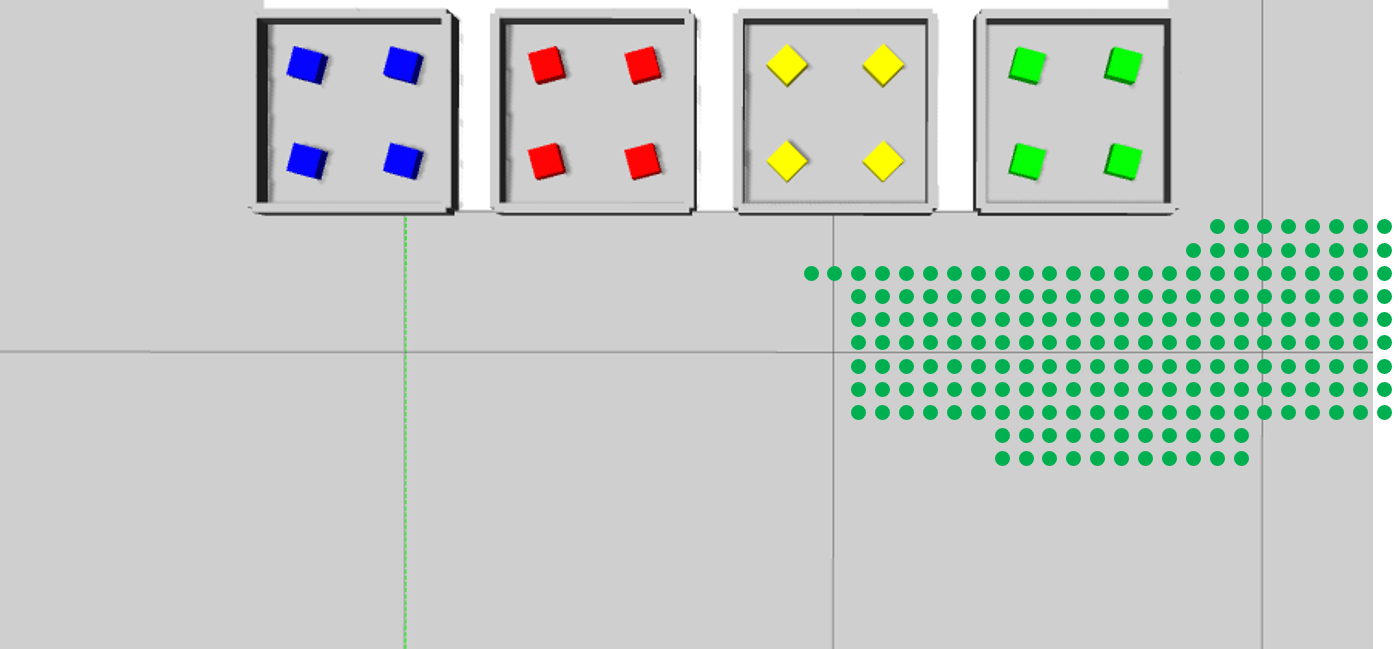}
    \end{subfigure}
    \caption{The base regions obtained by the proposed IK query method.}
    \label{fig:one_bin_result_IKRDB}
\end{figure}
\subsection{Intersection of Base Regions}
To reduce the number of base movements, the mobile manipulator had better move to the intersections where the mobile manipulator is able to pick up the objects in multiple trays. For the IK solver approach, there are 5 intersections for the obtained 4 base regions of 4 trays, all of them are the intersections of two base regions, while for the IK query approach, there are the intersections of two base regions and even the intersections of 3 base regions. The intersections and their associated trays are labeled with numbers in Fig. 8 and Fig. 9.
\begin{figure}
    \centering
    \includegraphics[width=0.46\textwidth]{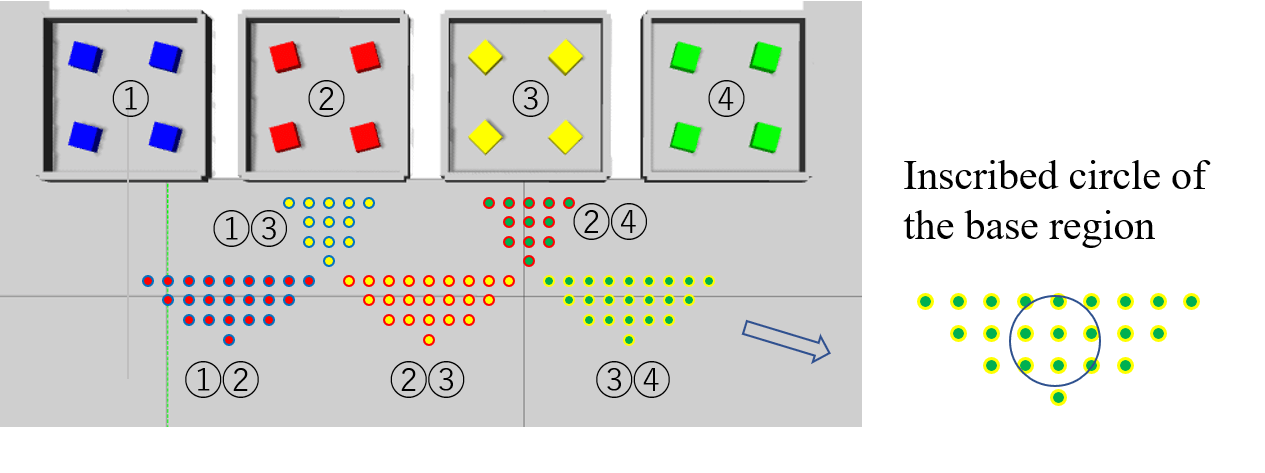}
    \caption{The intersections of the base regions in Fig. 6, the centers of their inscribed circles are the most robust base positions.}
    \label{fig:RDB_IK_4base_regions}
\end{figure}
\begin{figure}
    \centering
    \begin{subfigure}[h]{0.23\textwidth}
        \centering
        \includegraphics[width=1\textwidth]{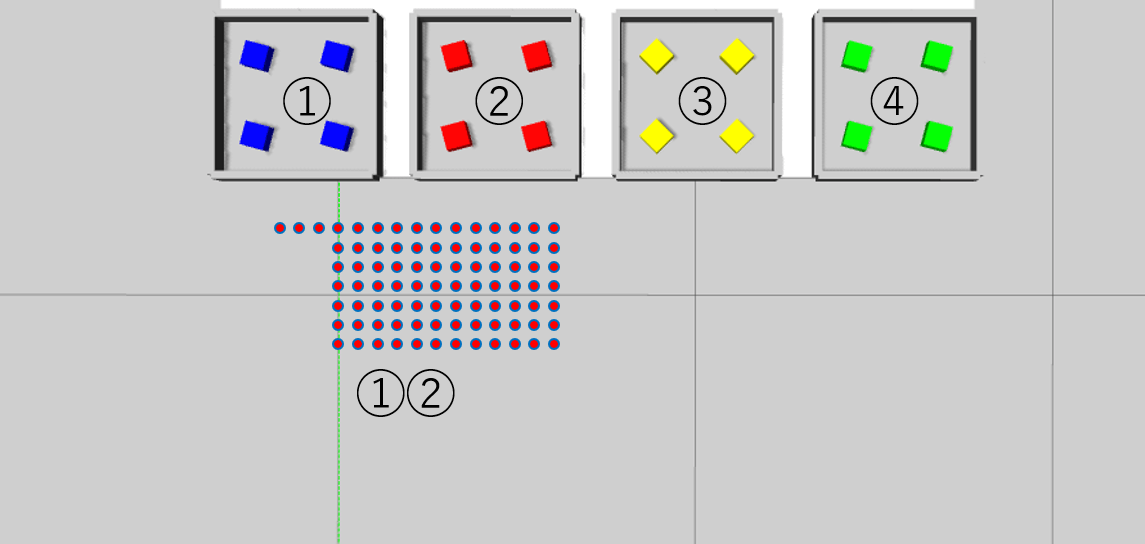}
        \caption{$P_{B1} \cap P_{B2}$}
    \end{subfigure}
    \begin{subfigure}[h]{0.23\textwidth}
        \centering
        \includegraphics[width=1\textwidth]{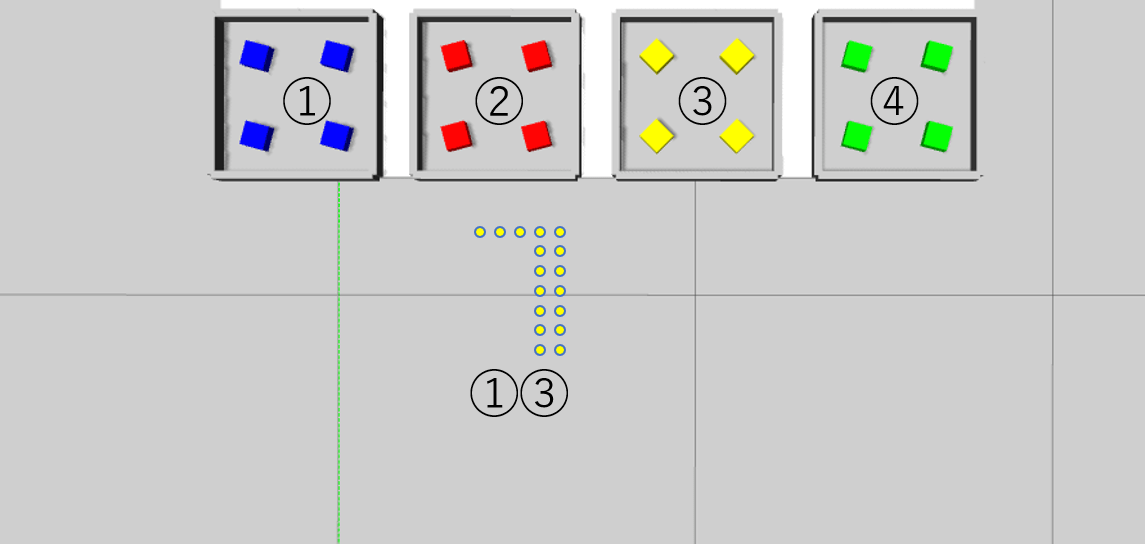}
        \caption{$P_{B1} \cap P_{B3}$}
    \end{subfigure}
    \begin{subfigure}[h]{0.23\textwidth}
        \centering
        \includegraphics[width=1\textwidth]{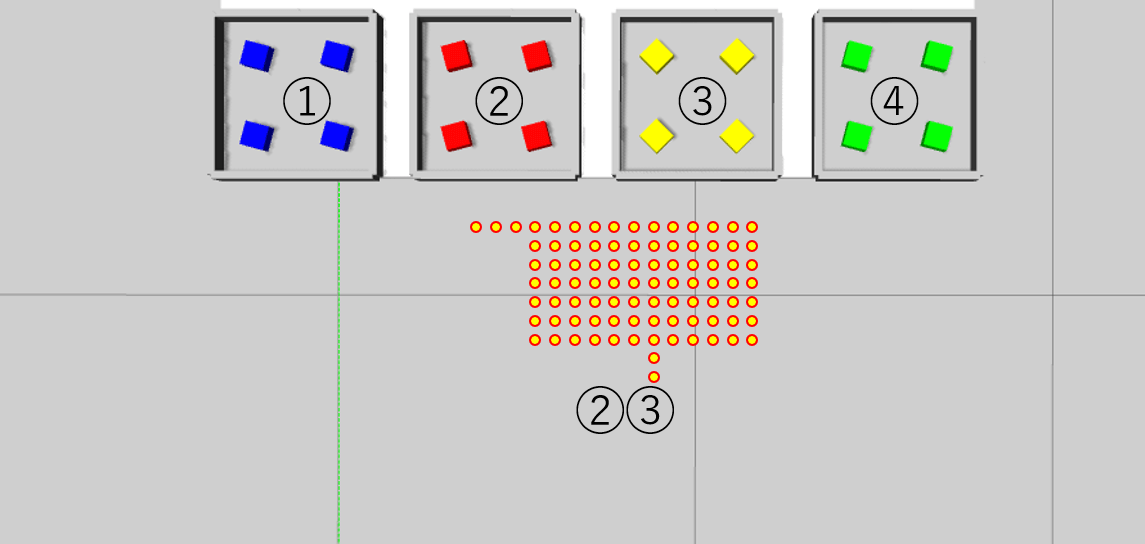}
        \caption{$P_{B2} \cap P_{B3}$}
    \end{subfigure}
    \begin{subfigure}[h]{0.23\textwidth}
        \centering
        \includegraphics[width=1\textwidth]{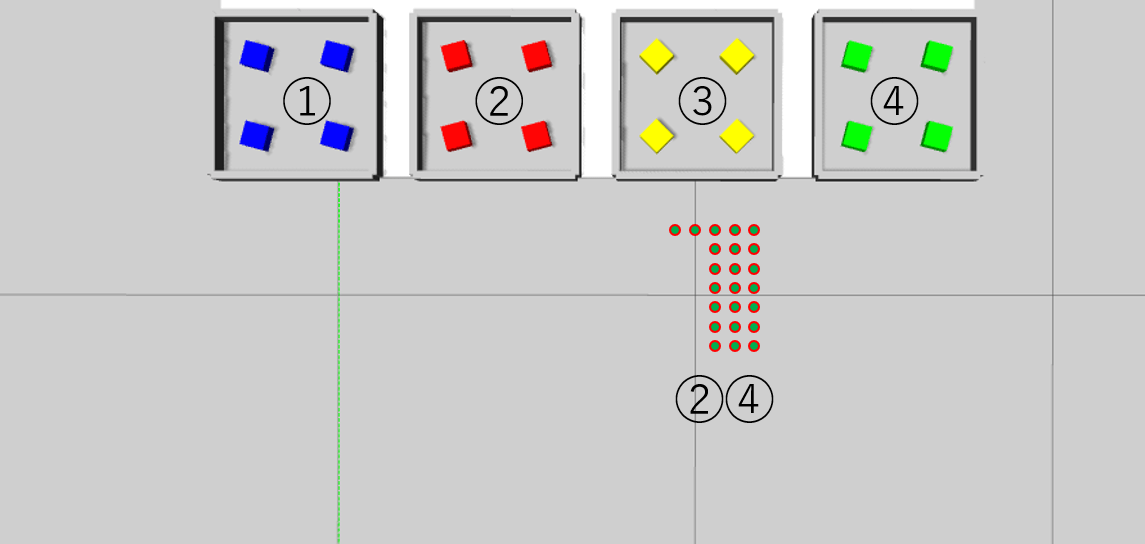}
        \caption{$P_{B2} \cap P_{B4}$}
    \end{subfigure}
    \begin{subfigure}[h]{0.23\textwidth}
        \centering
        \includegraphics[width=1\textwidth]{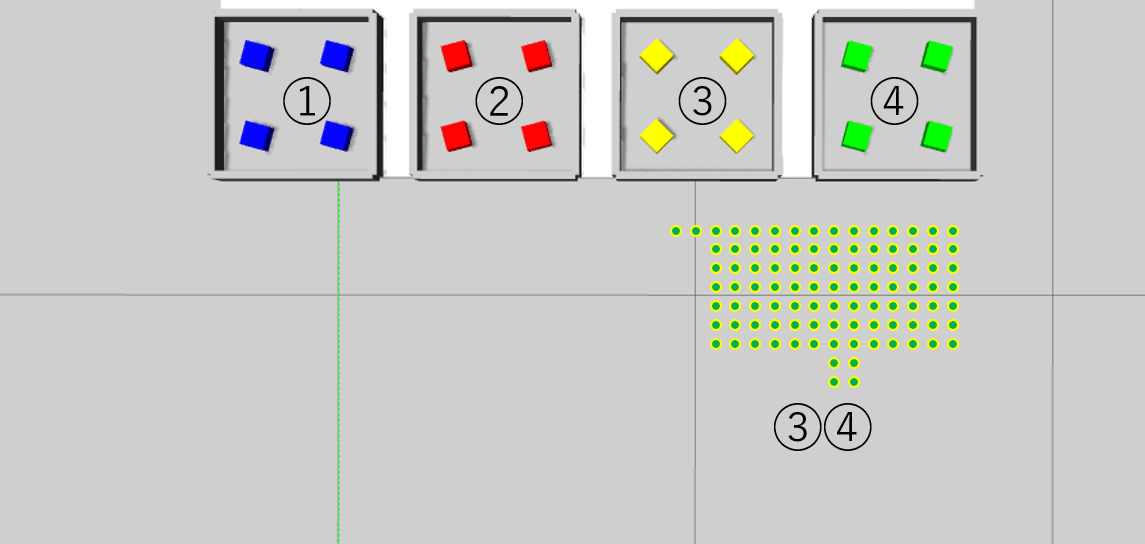}
        \caption{$P_{B3} \cap P_{B4}$}
    \end{subfigure}
    \begin{subfigure}[h]{0.23\textwidth}
        \centering
        \includegraphics[width=1\textwidth]{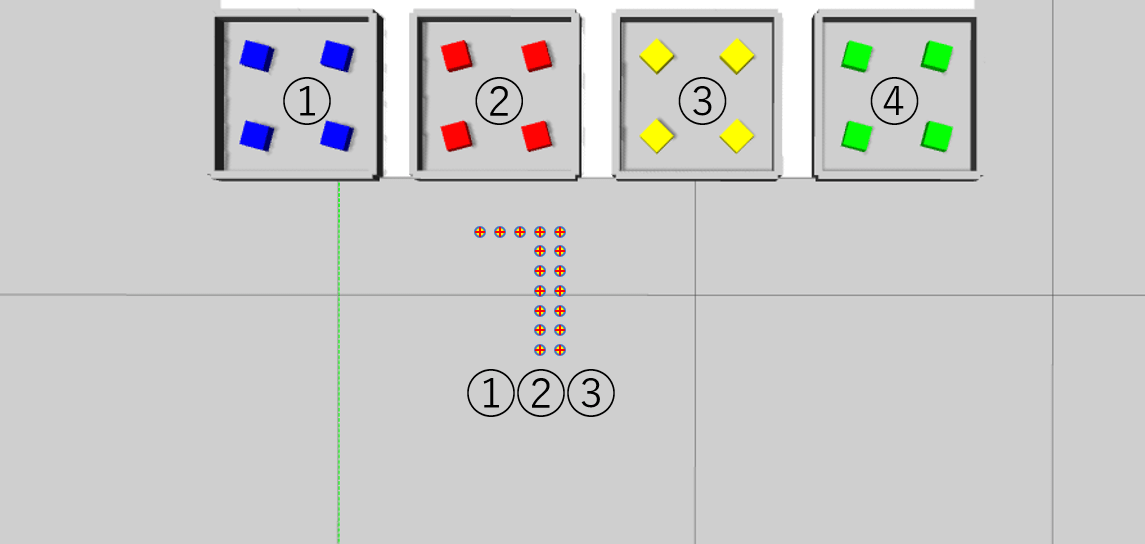}
        \caption{$P_{B1} \cap P_{B2} \cap P_{B3}$}
    \end{subfigure}
    \begin{subfigure}[h]{0.23\textwidth}
        \centering
        \includegraphics[width=1\textwidth]{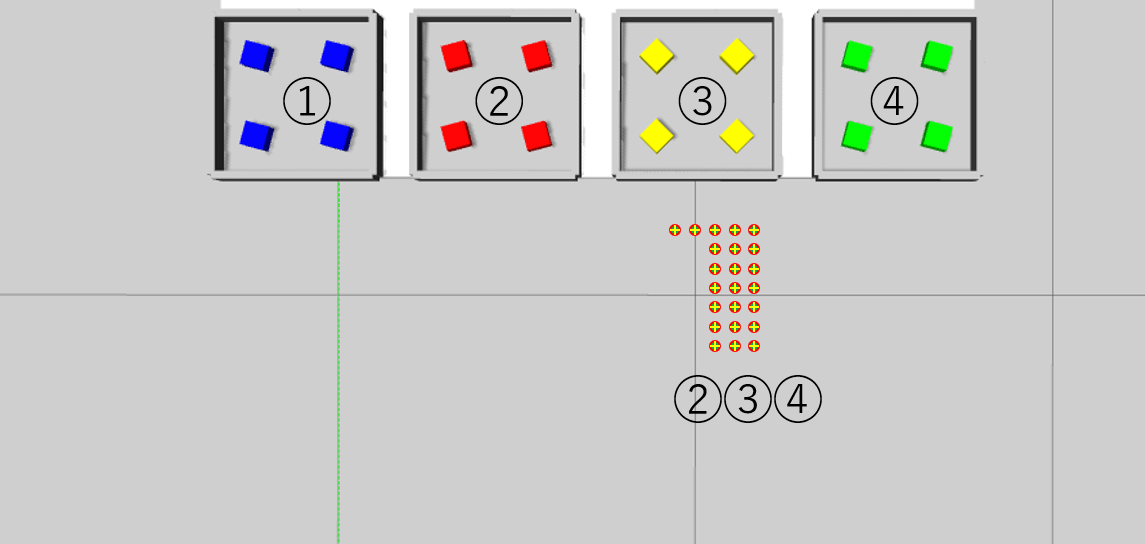}
        \caption{$P_{B2} \cap P_{B3} \cap P_{B4}$}
    \end{subfigure}
    \caption{All the intersections of the base regions in Fig. 7. (a)$\sim$(e) are all the intersections of two base regions, (f)$\sim$(g) are all the intersections of three base regions.}
    \label{fig:IKRDB_intersections}
\end{figure}
\subsection{Base Positioning Uncertainty}
Practically, the mobile manipulator is not able to arrive at exactly the same position as planned, the positioning uncertainty is governed by sensor accuracy and environmental complexity. It will be described in section \MakeUppercase{\romannumeral 6} that, a SLAM package is used to locate the mobile manipulator in the environment, the positioning error, as a result of numerous influencing factors, is assumed to be random and homogeneous in different directions. Let the average base positioning error be $\bar{\sigma}$(m), the mobile manipulator is most likely to arrive at a position $\bar{\sigma}$(m) away from the desired position. Therefore, in some base positions close to the boundary, the mobile manipulator may fail to reach the all the objects in the tray when positioning uncertainty is applied. The further the base position is from boundary, the more robust the point will be in terms of base positioning uncertainty. Therefore, the most robust base position is specified by the center of the inscribed circle of the base region or intersection (Fig. 8). If the radius of the inscribed circle of an intersection is smaller than the base positioning uncertainty level, the intersection is regarded as unreliable, and should not be applied to reduce the base sequence size. Notice that, the overall operation time is scarcely influenced by choosing different positions within the base region or intersection, due to their limited area, thus the robustness is given much higher priority in this stage without sacrificing much performance.

\subsection{Path Planning}
\begin{figure}
    \centering
    \begin{subfigure}[h]{0.23\textwidth}
        \centering
        \includegraphics[width=1\textwidth]{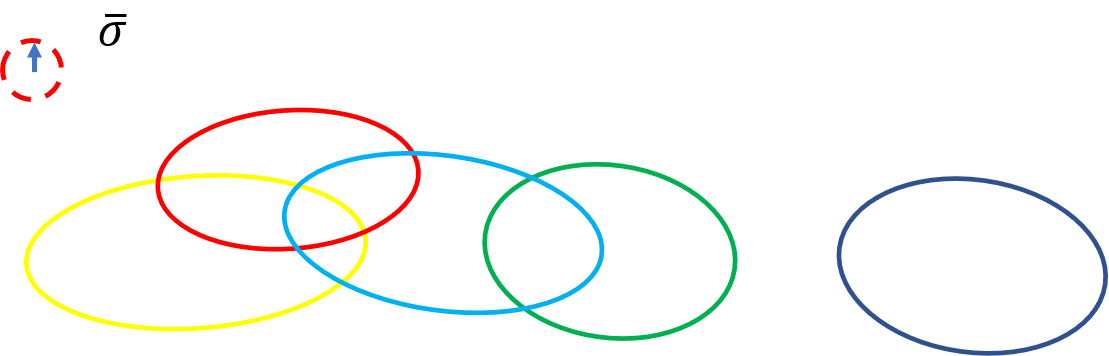}
        \caption{}
    \end{subfigure}
    \begin{subfigure}[h]{0.23\textwidth}
        \centering
        \includegraphics[width=1\textwidth]{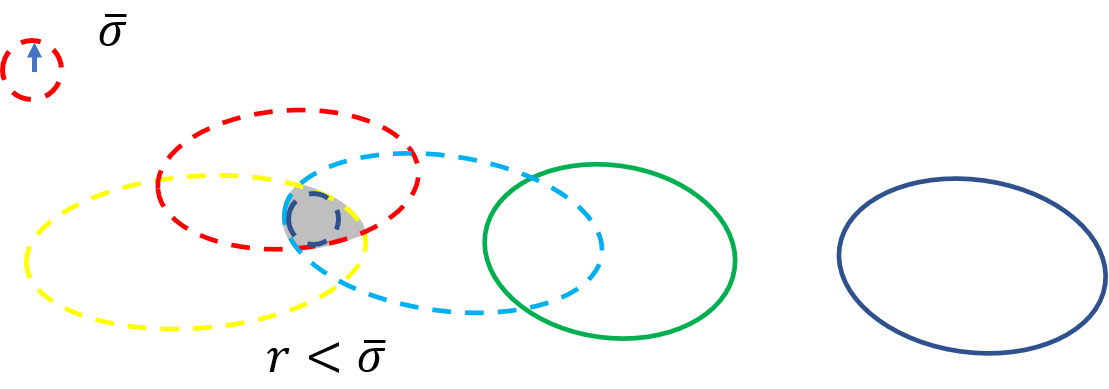}
        \caption{}
    \end{subfigure}
    \begin{subfigure}[h]{0.23\textwidth}
        \centering
        \includegraphics[width=1\textwidth]{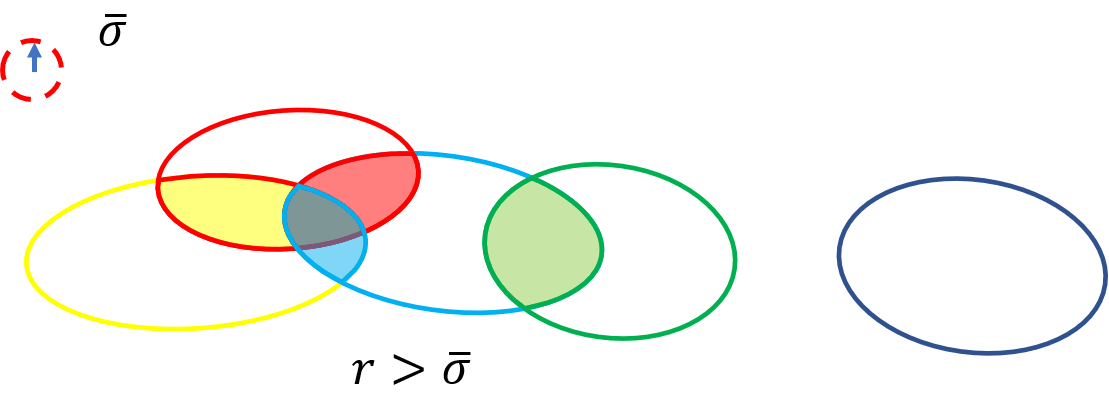}
        \caption{}
    \end{subfigure}
     \begin{subfigure}[h]{0.23\textwidth}
        \centering
        \includegraphics[width=1\textwidth]{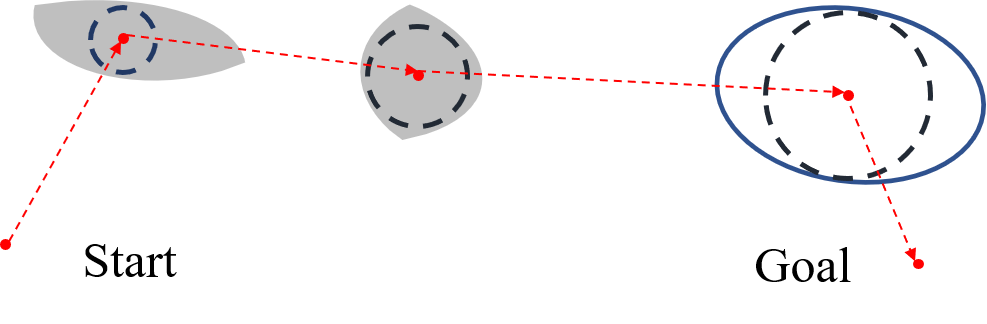}
        \caption{}
    \end{subfigure}
    \caption{The procedure of path planning: (a) From left to right are five base regions $P_{B1} \sim P_{B5}$ for 5 trays. (b) ${\it InscribedRadius}(P_{B1} \cap P_{B2} \cap P_{B3})<\bar{\sigma}$, thus discarded. (c) Four feasible second order intersections are filled with different colors. (d) The planned intersections $P_{B1} \cap P_{B2}$, $P_{B3} \cap P_{B4}$ and $P_{B5}$ are connected by the shortest path.}
    \label{fig:path_planning}
\end{figure}
Given an assembly task to collect the objects in trays \{$tray_1, tray_2,\dots,tray_n$\},  a set of base regions \{$P_1, P_2, \dots, P_n$\} and their intersections can be obtained following the proposed method. Let $\cap_{i}^{k}P_{Bi}$, $(1 \leq i \leq n)$, denote all the k-th order intersections, which are the intersections of k base regions, base regions are regarded as first order intersections, and $\lambda$ be largest k, then $\{\cap_{i}^{1}P_{Bi},\cap_{i}^{2}P_{Bi},\dots,\cap_{i}^{\lambda}P_{Bi}\}$ is the set of all the possible intersections. While considering the base positioning uncertainty, the intersection are removed from the set if its inscribed circle is smaller than $\bar{\sigma}$. Then from the remaining set of size $N$, we select $m$ intersections that cover all the target tray and $m$ is minimized as possible, this is equivalent to the problem of assigning $(N-m)$ zeros and $m$ ones to base sequence vector $[x_1,x_2,\dots,x_N]^T$ and minimizing $\sum_{i=1}^N x_i$, where $x_i=\{0,1\}$, subject to: $\sum_{i=1}^N a_{1i}x_i=1,\sum_{i=1}^N a_{2i}x_i=1\dots,\sum_{i=1}^N a_{ni}x_i=1$, where $a_{si}, s=\{1,2,\dots,n\}$, is 1 if $P_{Bs}$ is reached by the intersection, and $x_i=1$ if the robot moves to the corresponding intersection. This is the 0-1 knapsack problem and can be solved by BB method \cite{kolesar1967}. Finally, take the centers of these $m$ intersections and connect them, together with starting and goal positions, by the shortest path.

For example, in Fig. 10, there are 5 base regions of 5 trays, 1 third order intersection and 4 second order intersections, while the third order intersection $P_{B1} \cap P_{B2} \cap P_{B3} < \bar{\sigma}$, thus removed from the total set of intersections. From the remaining 9 intersections, three of them are planned to reach all of the trays, then the shortest path can be searched, that connects the starting and goal position, via the the centers of their inscribed circles. If the sequence size becomes too large for searching, the shortest path can be approximated by SA method \cite{cerny1985}.

\section{Experimental Investigation}
\begin{figure}
    \centering
    \begin{subfigure}[h]{0.23\textwidth}
        \centering
        \includegraphics[width=\textwidth]{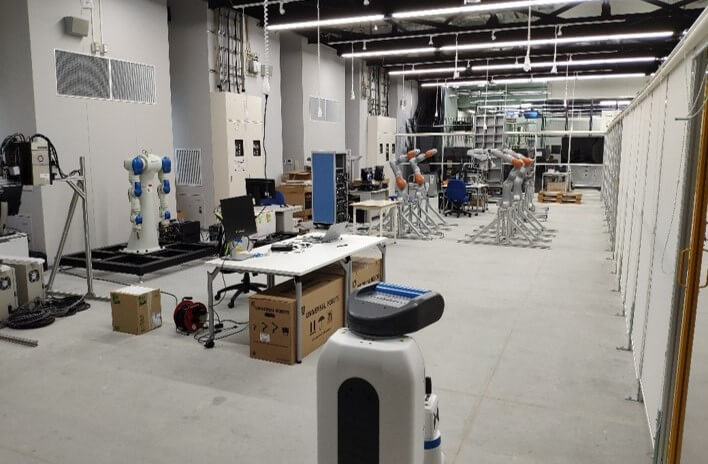}
    \end{subfigure}
    \begin{subfigure}[h]{0.23\textwidth}
        \centering
        \includegraphics[width=\textwidth]{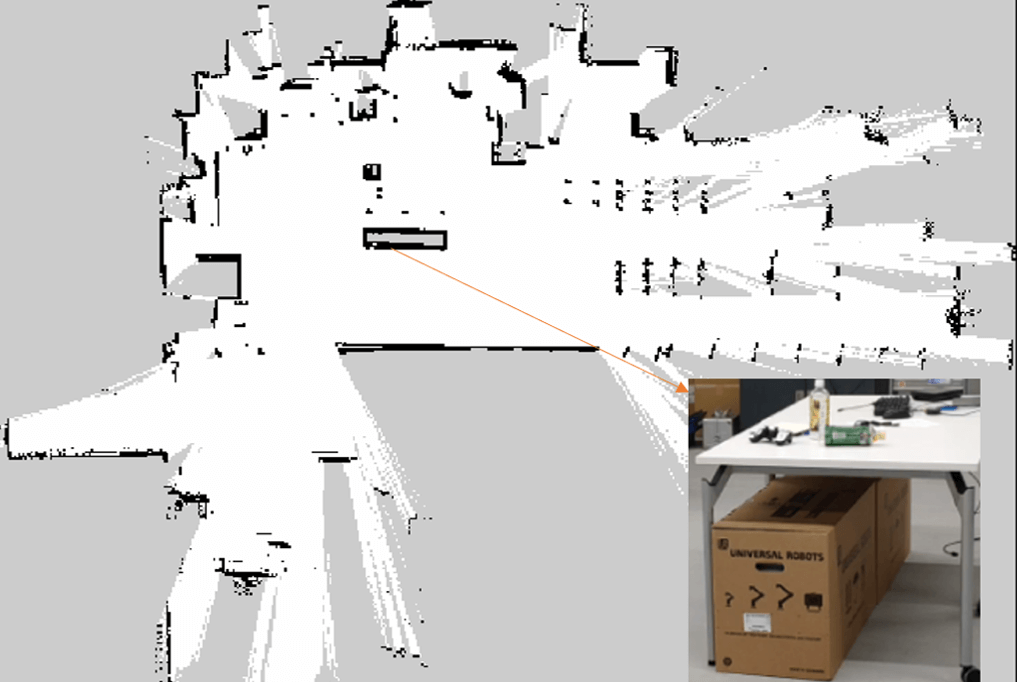}
    \end{subfigure}
    \begin{subfigure}[h]{0.16\textwidth}
    \centering
    \includegraphics[width=\textwidth]{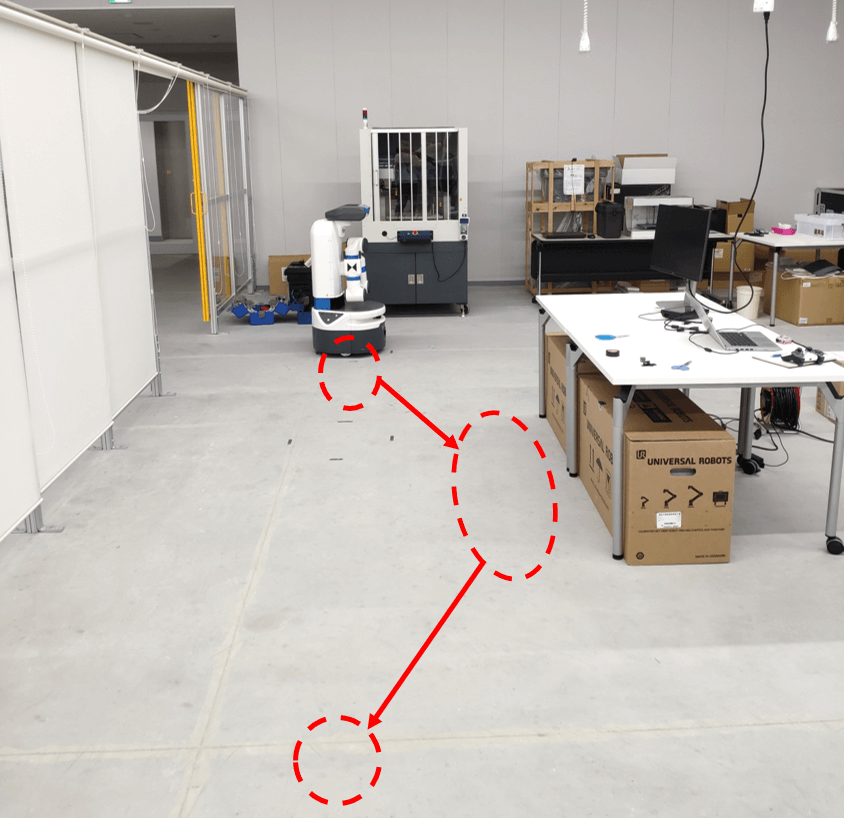}
    \end{subfigure}
    \begin{subfigure}[h]{0.30\textwidth}
    \centering
    \includegraphics[width=\textwidth]{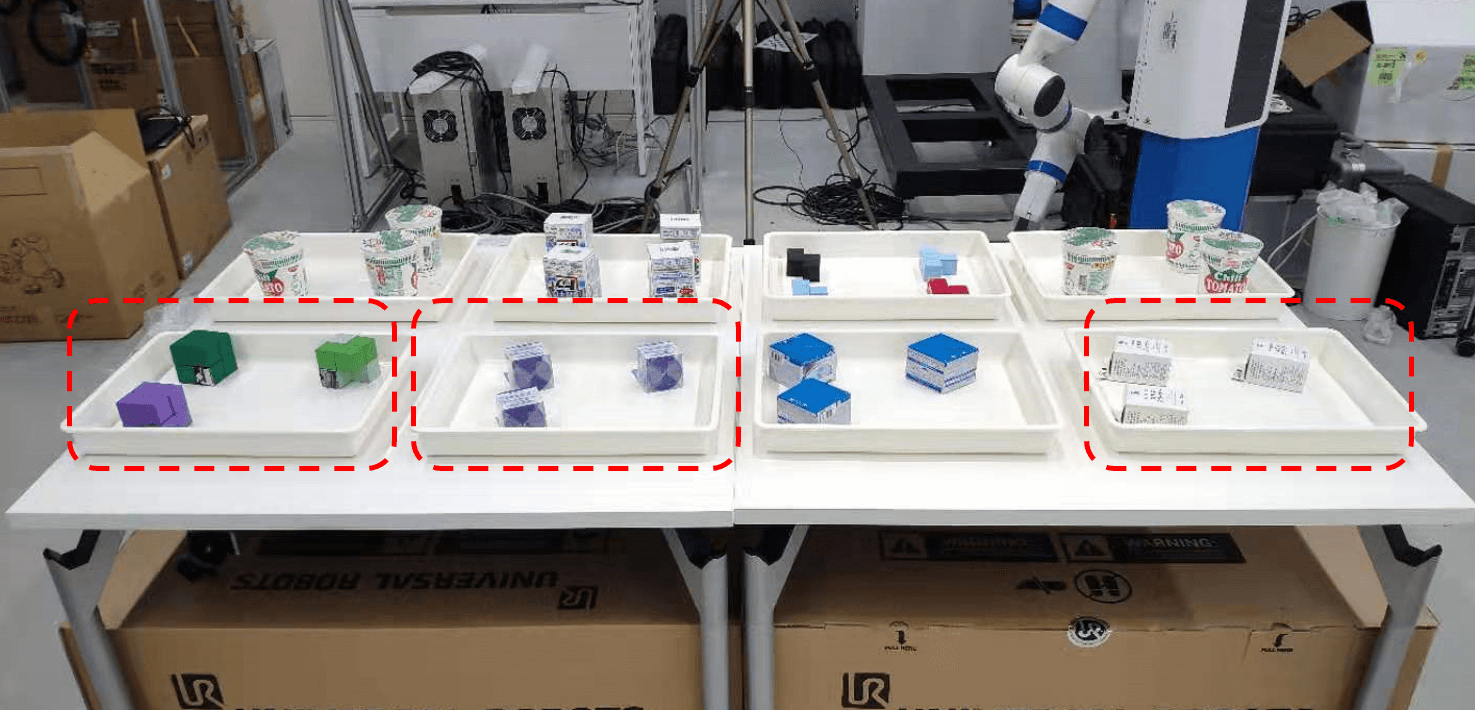}
    \end{subfigure}
    \caption{Experiment setup: (Top left) Indoor experimental environment. (Top right) A 2d map build by the laser scanner. (Bottom left) Task overview. (Bottom right) Target objects and their poses in three trays.}
    \label{fig:experiment_setup}
\end{figure}
Experiments are performed to valid our method in practical application. The Fetch robot \cite{fetch}, a single arm mobile manipulator, is used in this study. The robot navigates in an indoor environment as shown in Fig. 11, it starts from the origin and picks up 3 objects stored in 3 different trays, then carries them to the goal position. 

\begin{figure}
    \centering
    \begin{subfigure}[h]{0.15\textwidth}
    \centering
    \includegraphics[width=\textwidth]{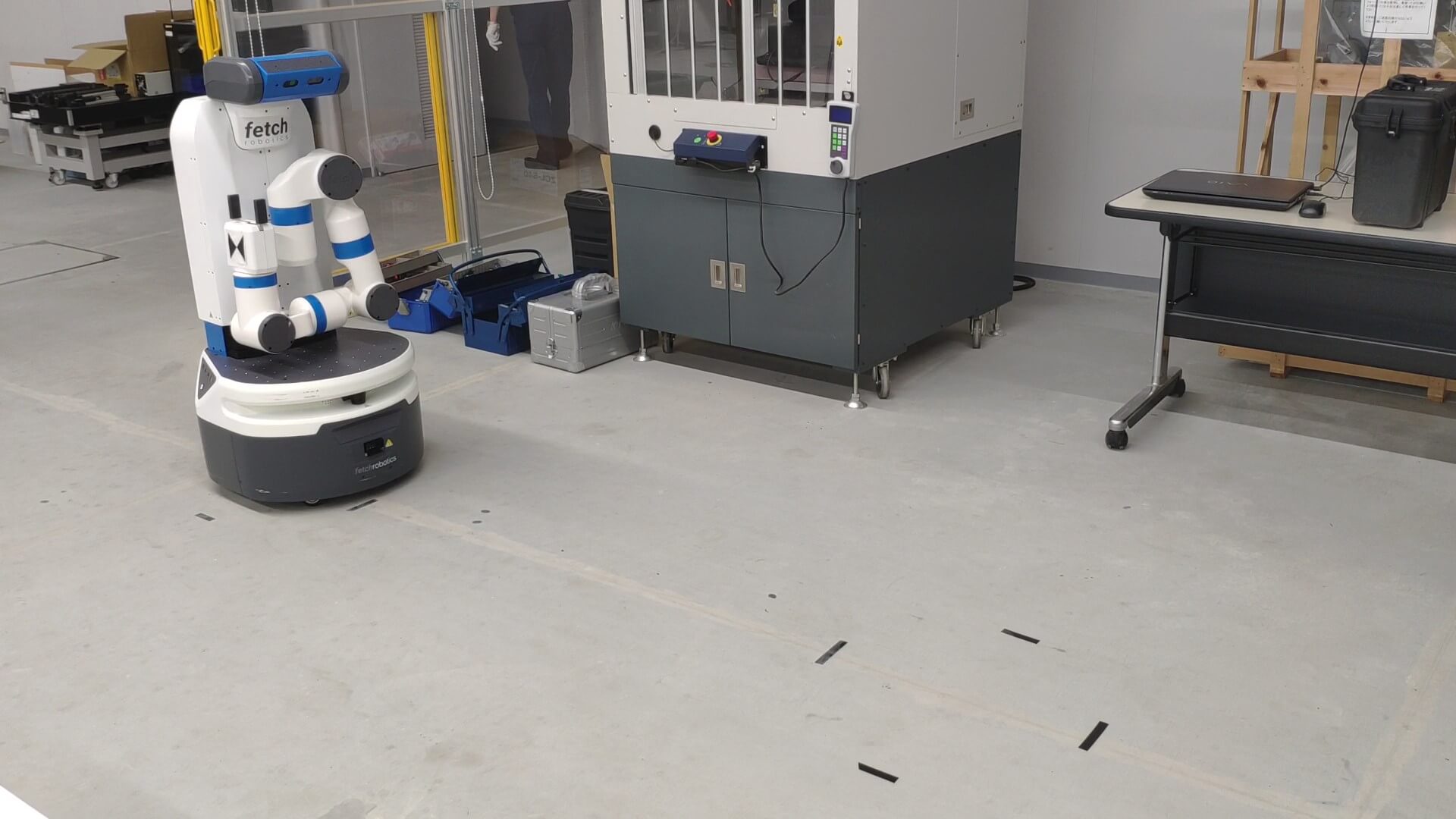}
    \end{subfigure}
    \begin{subfigure}[h]{0.15\textwidth}
    \centering
    \includegraphics[width=\textwidth]{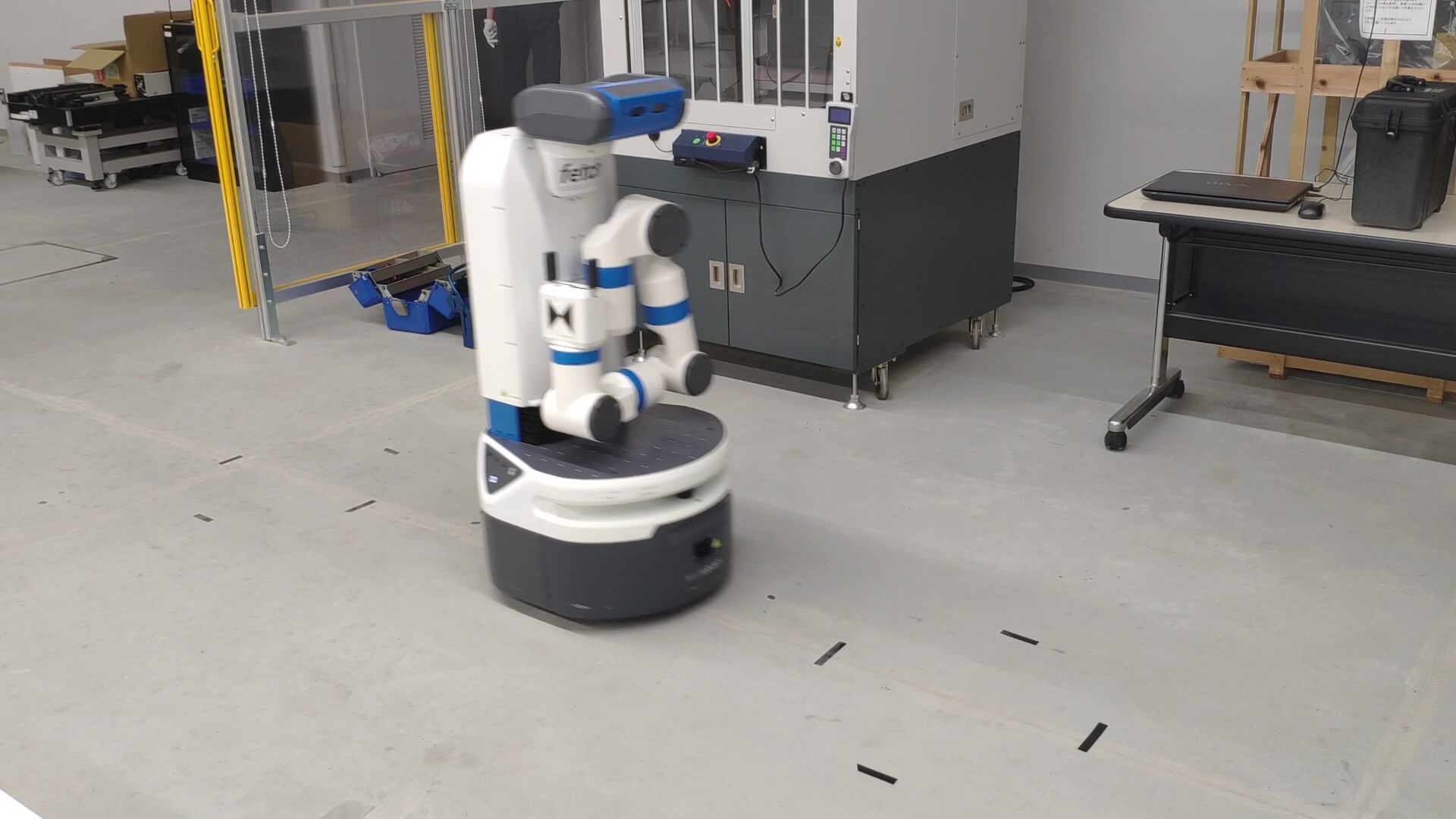}
    \end{subfigure}
    \begin{subfigure}[h]{0.15\textwidth}
    \centering
    \includegraphics[width=\textwidth]{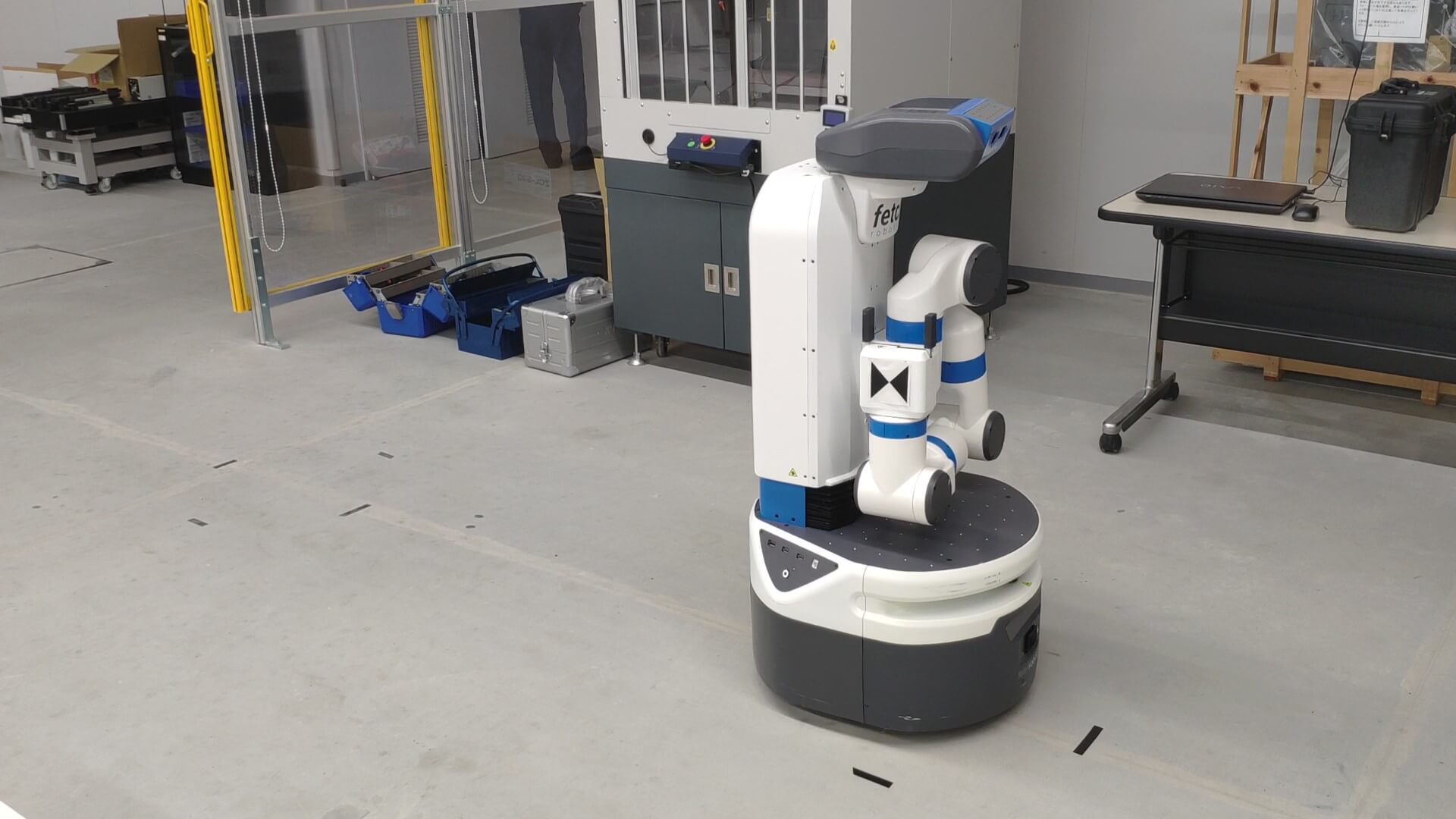}
    \end{subfigure}
    \caption{Test the base positioning accuracy and repeatability by looping the mobile manipulator between two fixed positions, the black tape on the ground marks the profile of mobile base in these two positions.}
    \label{fig:uncertainty}
\end{figure}
The base regions and intersections are calculated by the proposed method, and the results are already presented in Fig. 7 and Fig. 9. In order to obtain a robust base sequence, the base positioning uncertainty and repeatability are tested by looping the mobile manipulator between two fixed positions, as shown in Fig. 12, the actually arrived positions are observed to deviate about 10 cm from the desired positions. The base positioning error is the result of map accuracy, sensor accuracy, environmental complexity and the performance of the mechanical system, with so many factors involved, it is assumed to be random and homogeneous in any direction, therefore, the base positioning uncertainty level $\bar{\sigma}$ is set as 10 cm, then a sequence of base positions can be planned by algorithm described in section \MakeUppercase{\romannumeral 5}.D. As a result, the mobile manipulator should move to the center of $P_{B1} \cap P_{B2}$ and $P_{B4}$ to collect all the required parts.

\begin{figure}
    \centering
    \begin{subfigure}[h]{0.15\textwidth}
    \centering
    \includegraphics[width=\textwidth]{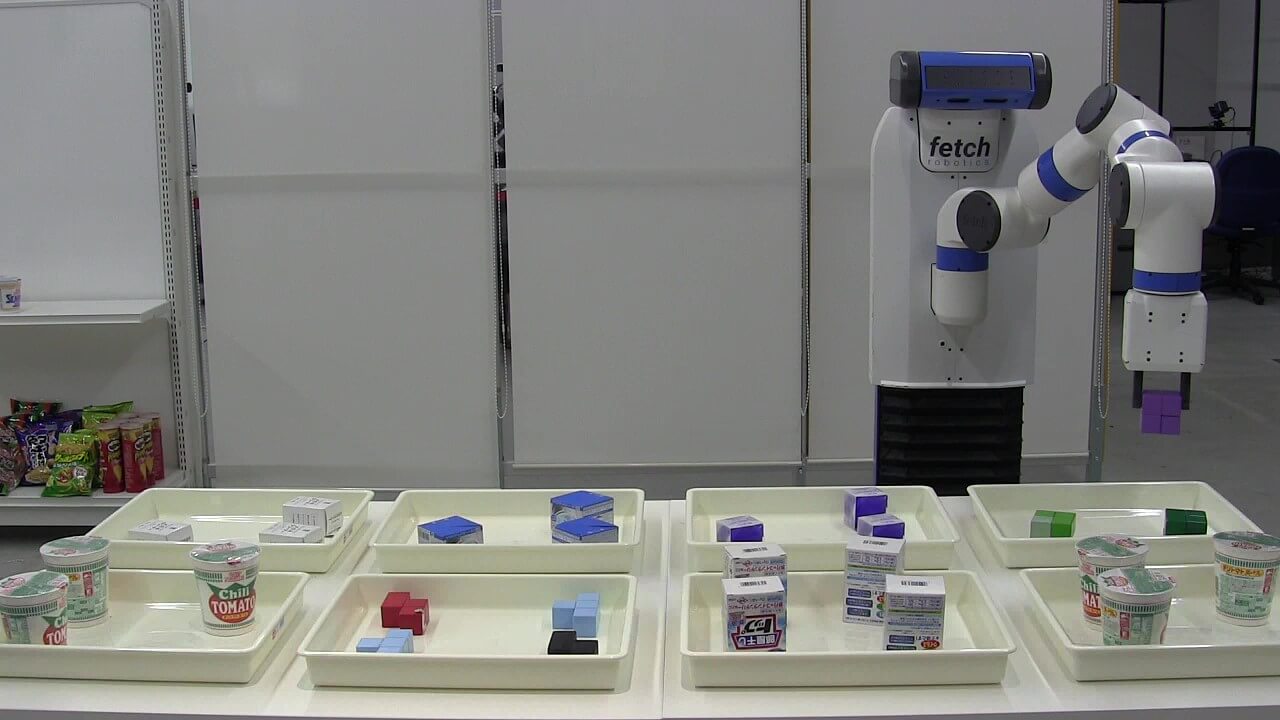}
    \end{subfigure}
    \begin{subfigure}[h]{0.15\textwidth}
    \centering
    \includegraphics[width=\textwidth]{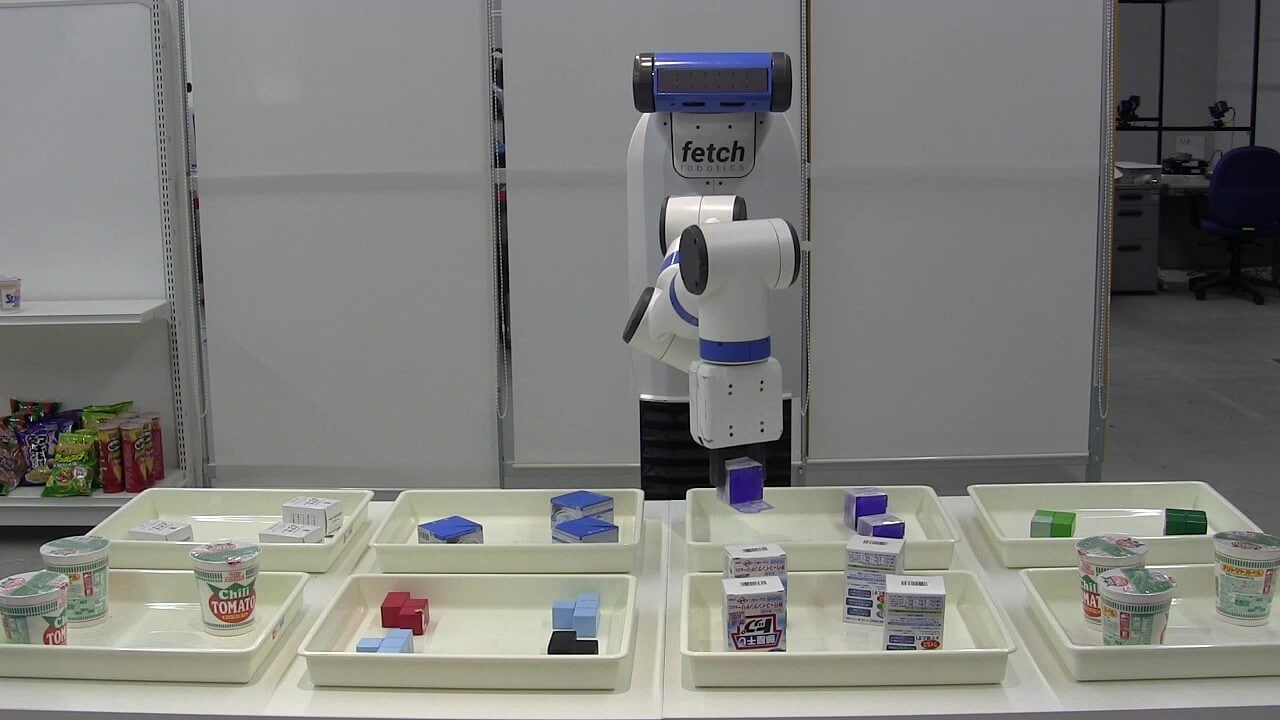}
    \end{subfigure}
    \begin{subfigure}[h]{0.15\textwidth}
    \centering
    \includegraphics[width=\textwidth]{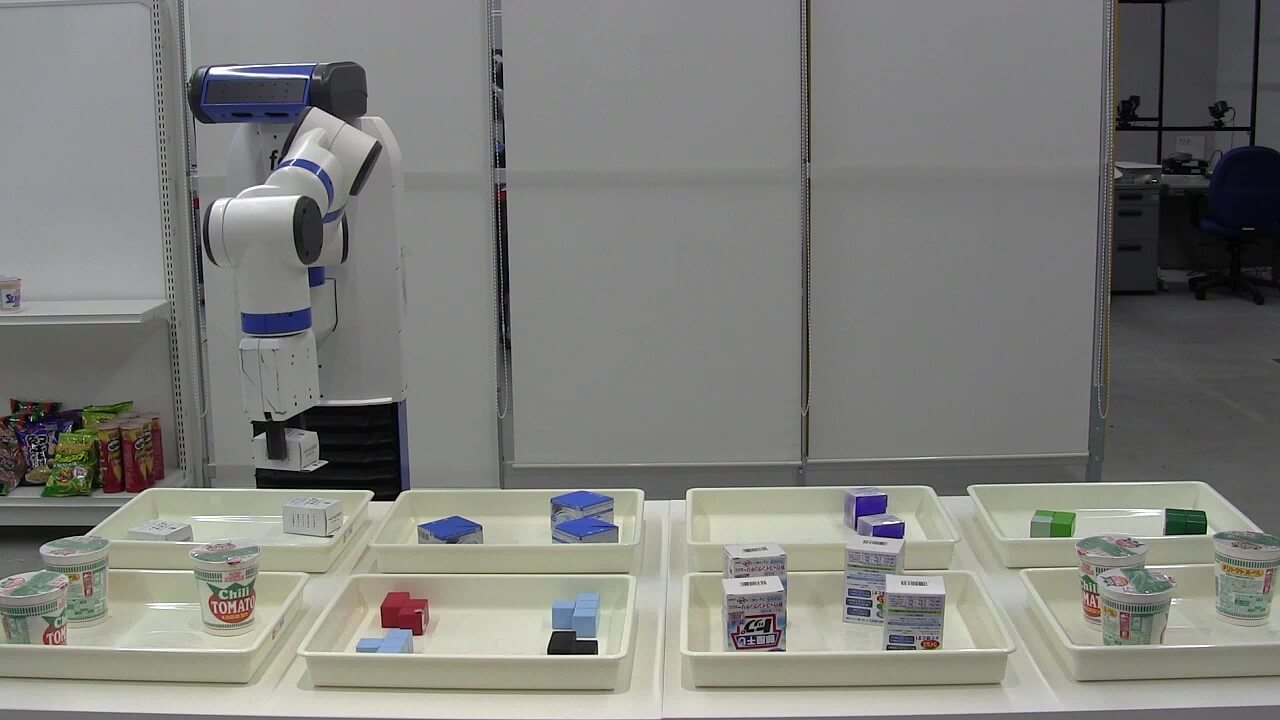}
    \end{subfigure}
    \begin{subfigure}[h]{0.15\textwidth}
    \centering
    \includegraphics[width=\textwidth]{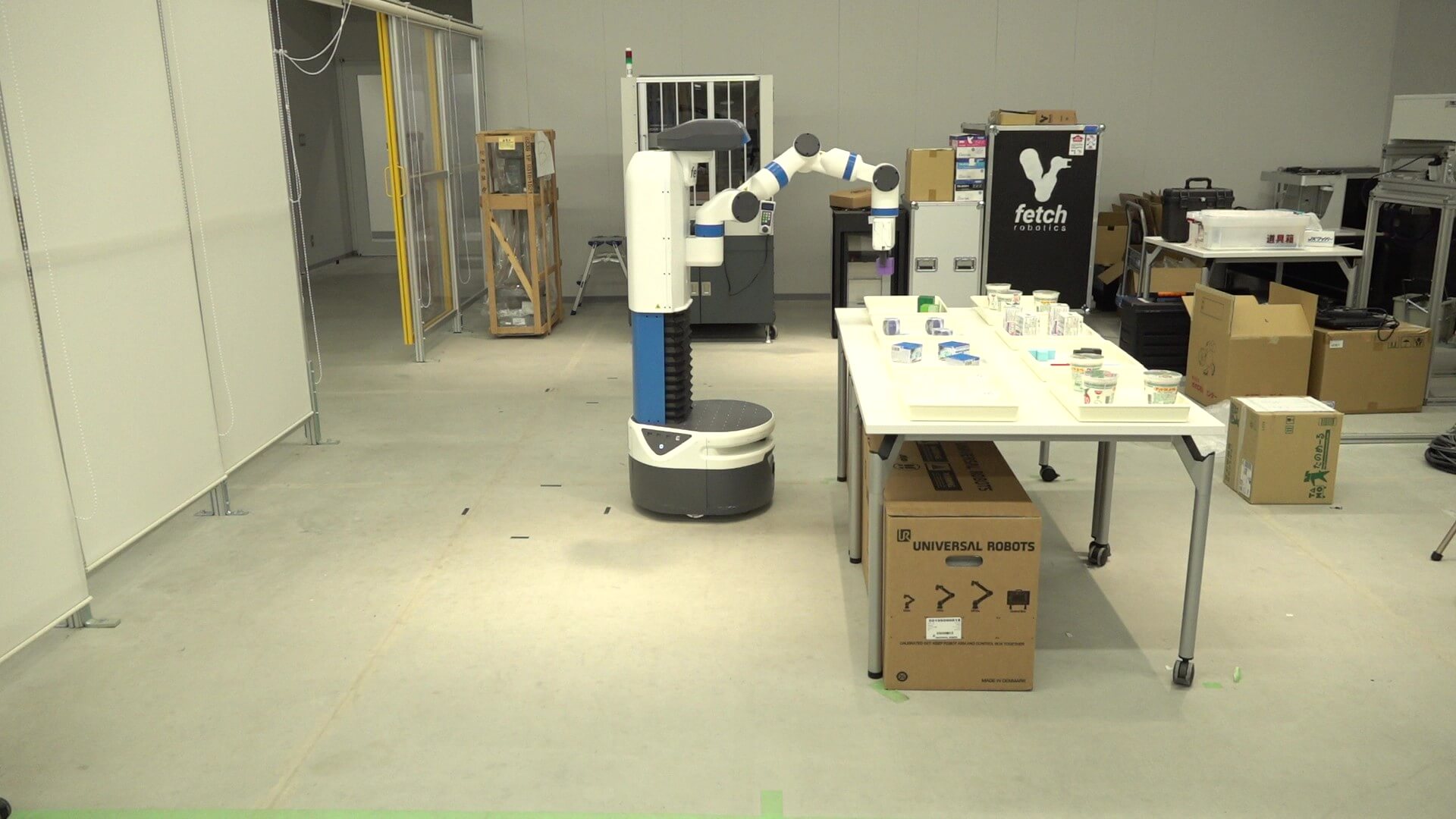}
    \end{subfigure}
    \begin{subfigure}[h]{0.15\textwidth}
    \centering
    \includegraphics[width=\textwidth]{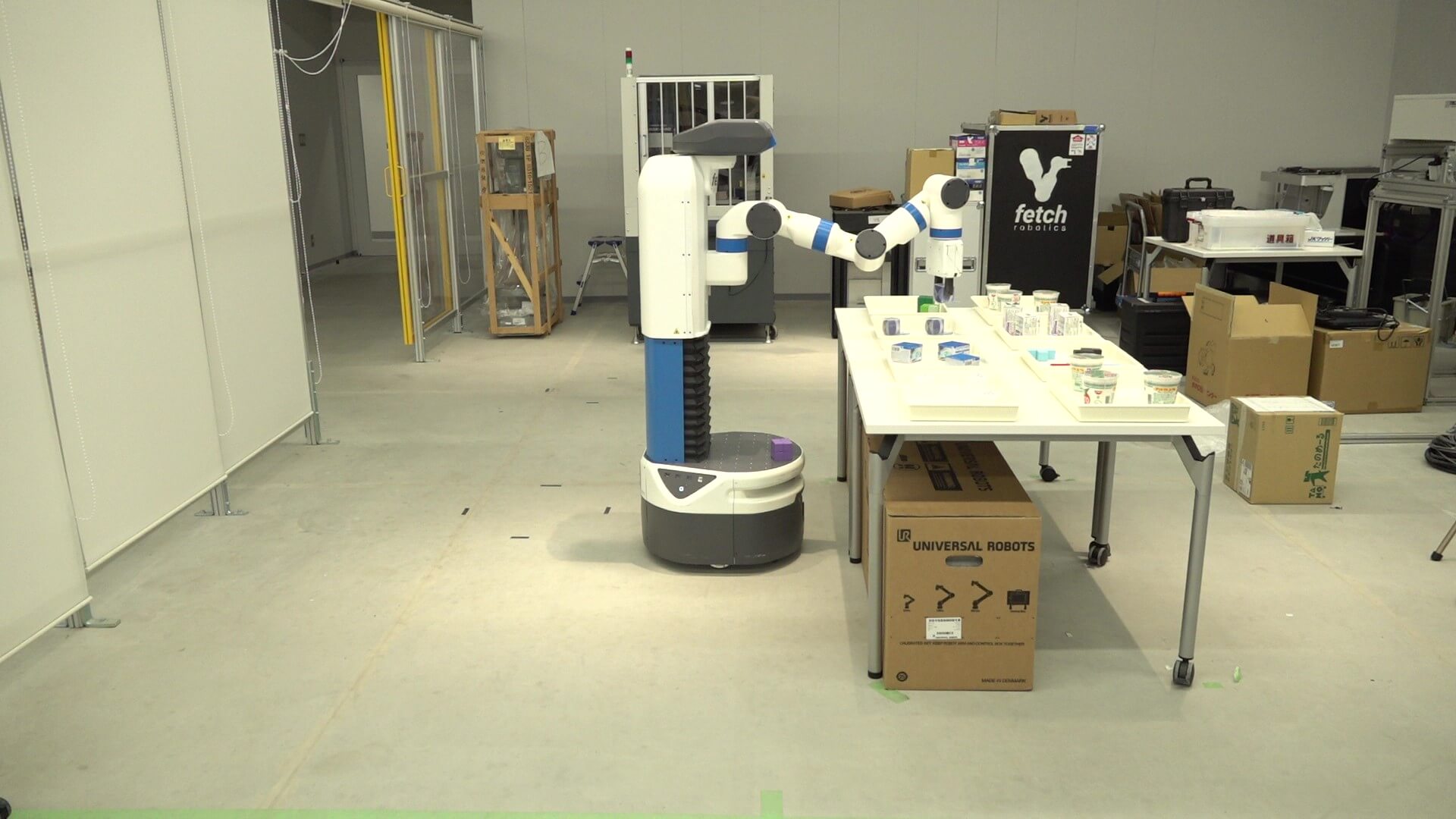}
    \end{subfigure}
    \begin{subfigure}[h]{0.15\textwidth}
    \centering
    \includegraphics[width=\textwidth]{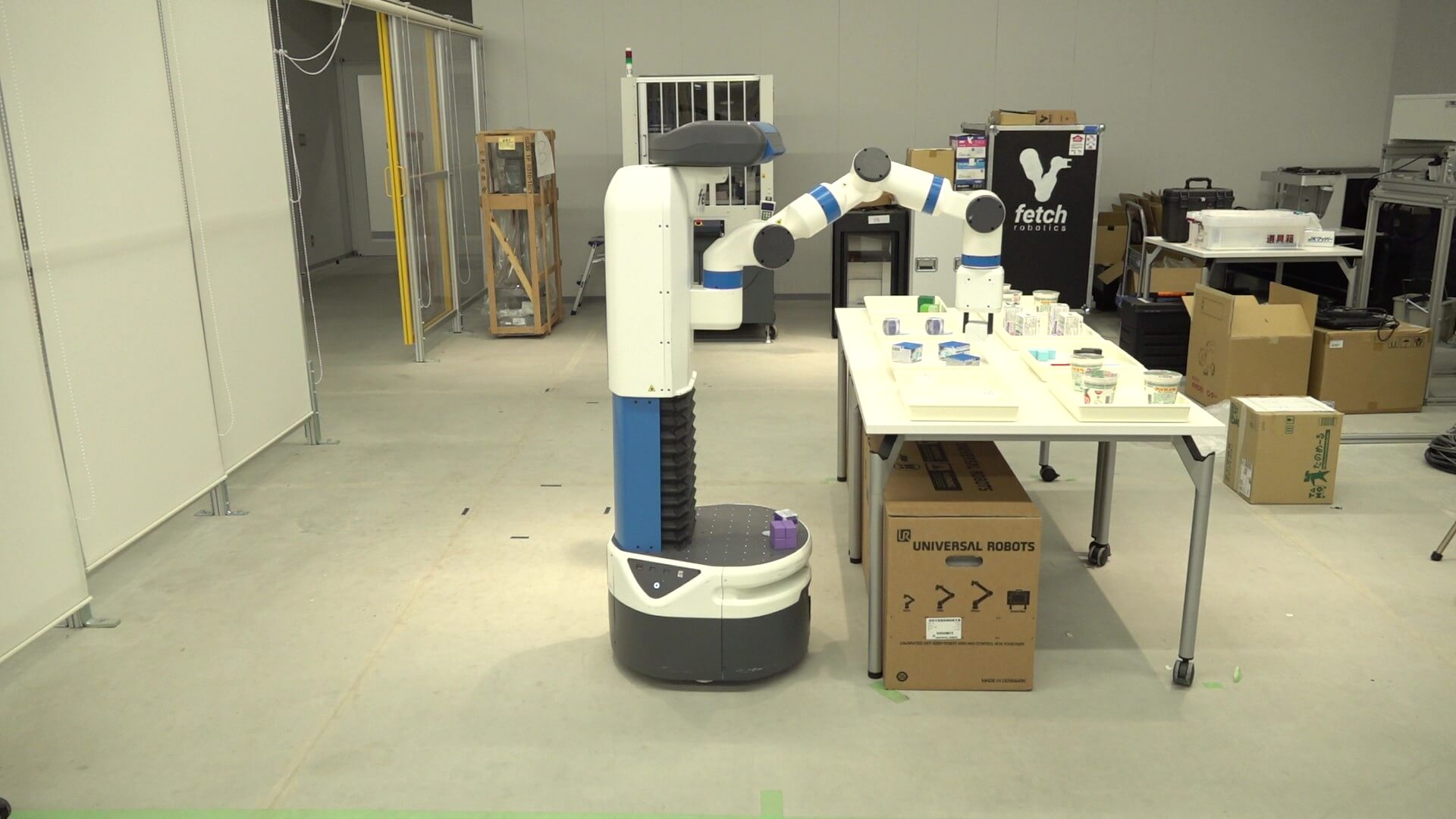}
    \end{subfigure}
    \caption{(Left) Move to $P_{B1}$ to pick up a part from $tray_1$. (Middle) Move to $P_{B2}$ to pick up a part from $tray_2$. (Right) Move to $P_{B4}$ to pick up a part from $tray_4$.}
    \label{fig:exper1}
\end{figure}
\begin{figure}
    \centering
    \begin{subfigure}[h]{0.15\textwidth}
    \centering
    \includegraphics[width=\textwidth]{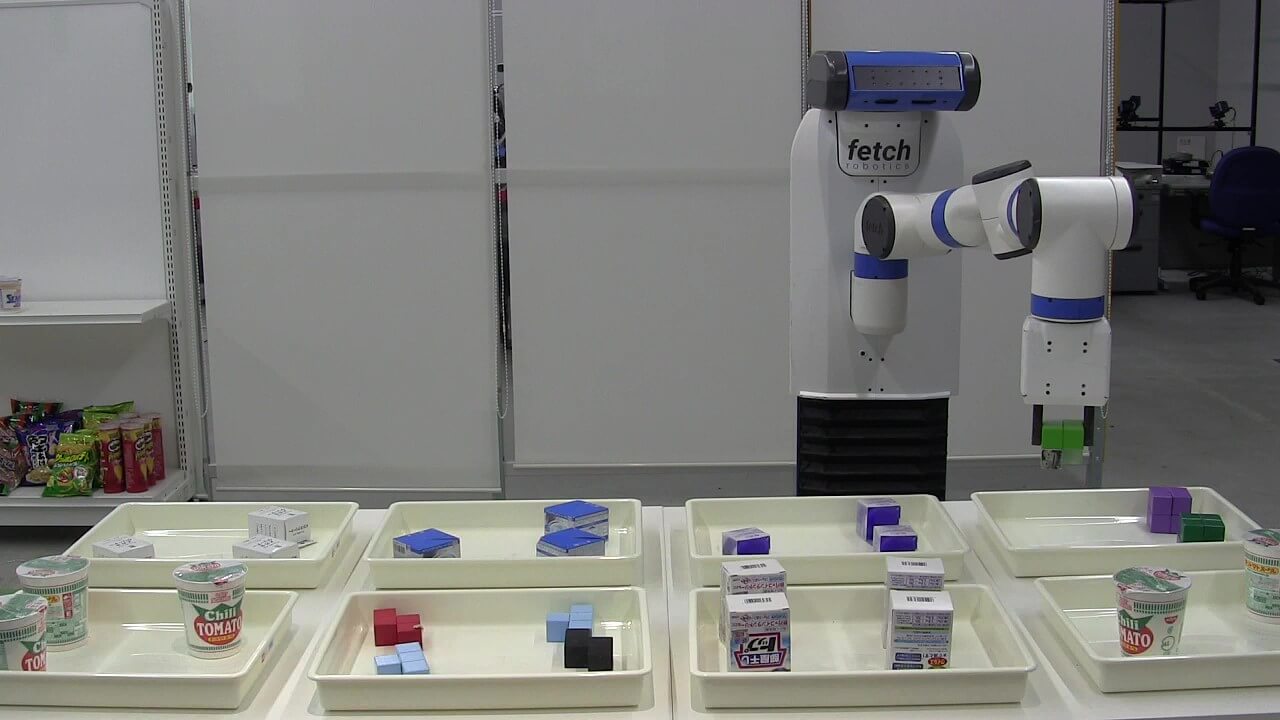}
    \end{subfigure}
    \begin{subfigure}[h]{0.15\textwidth}
    \centering
    \includegraphics[width=\textwidth]{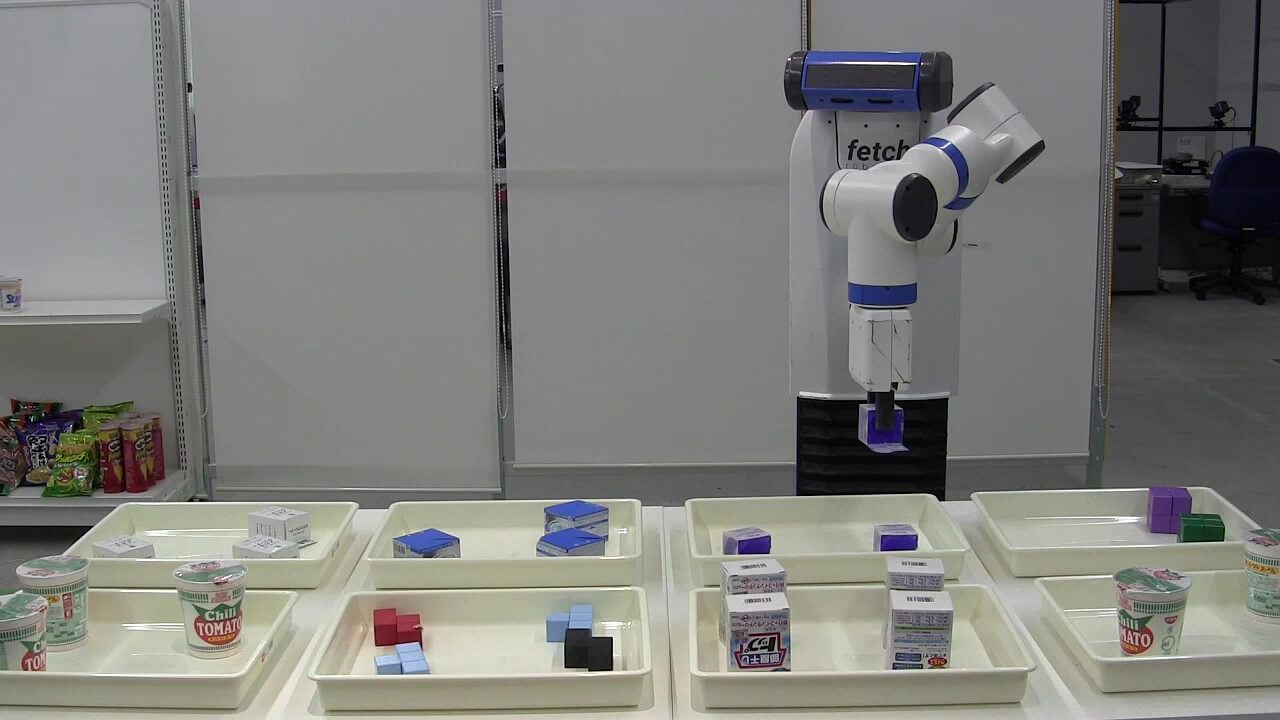}
    \end{subfigure}
    \begin{subfigure}[h]{0.15\textwidth}
    \centering
    \includegraphics[width=\textwidth]{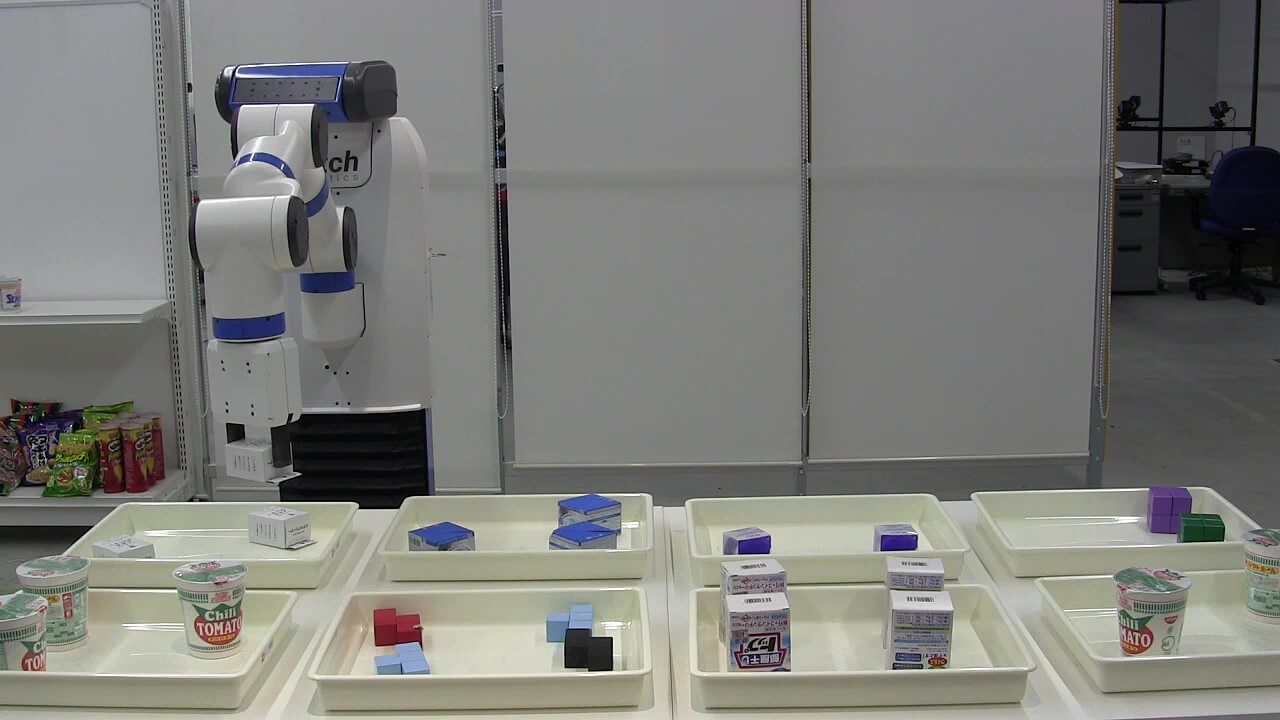}
    \end{subfigure}
    \begin{subfigure}[h]{0.15\textwidth}
    \centering
    \includegraphics[width=\textwidth]{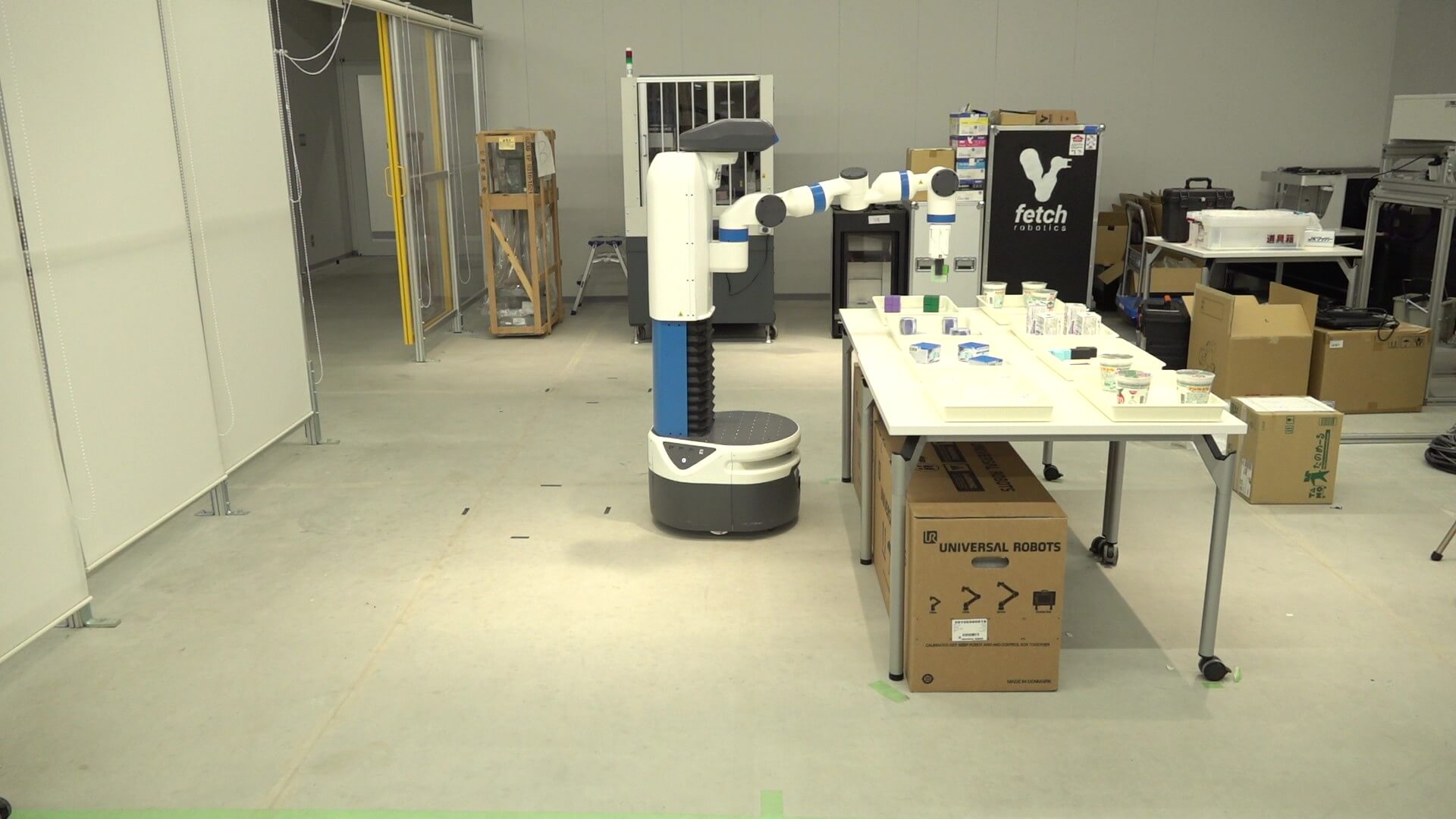}
    \end{subfigure}
    \begin{subfigure}[h]{0.15\textwidth}
    \centering
    \includegraphics[width=\textwidth]{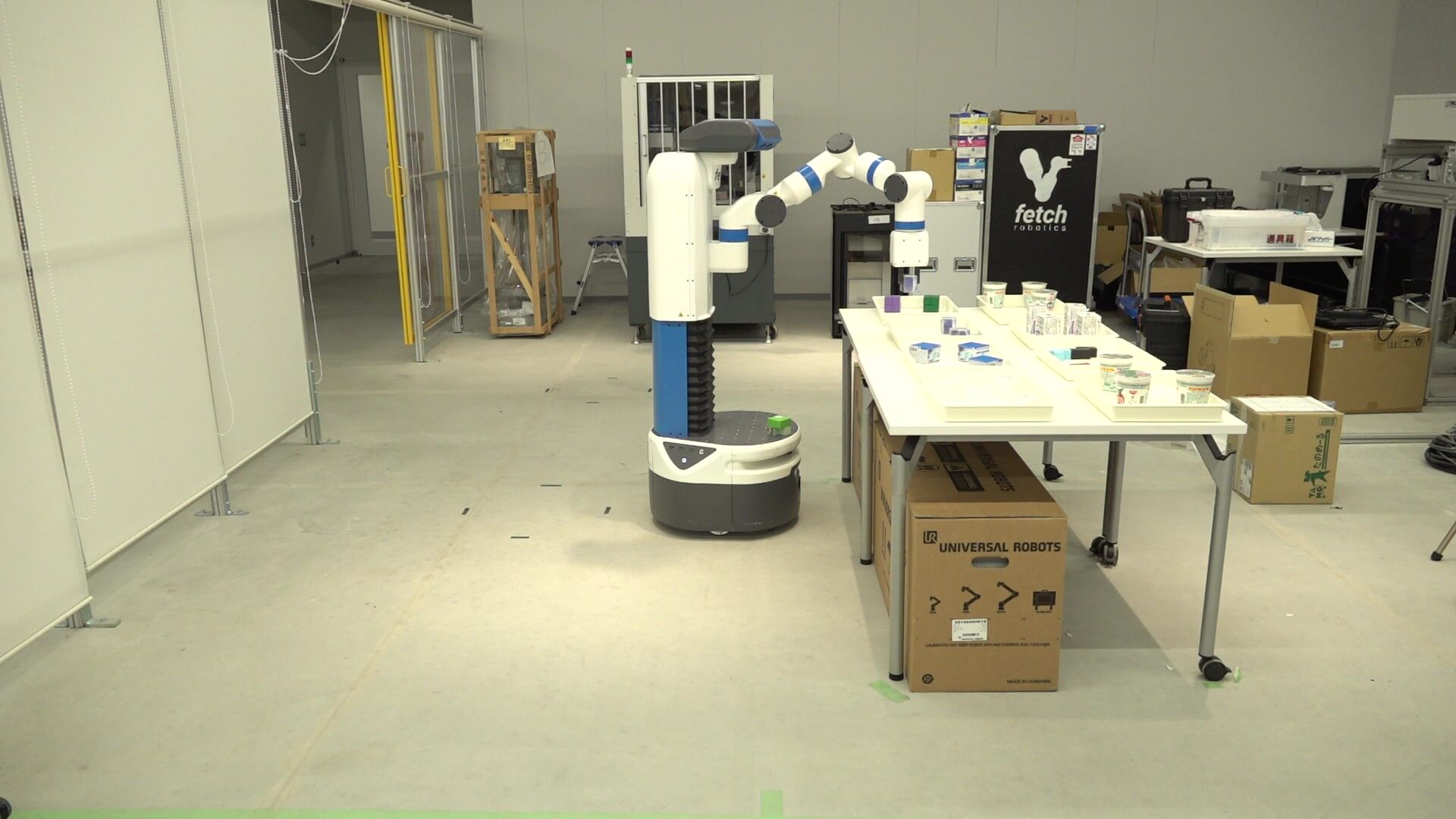}
    \end{subfigure}
    \begin{subfigure}[h]{0.15\textwidth}
    \centering
    \includegraphics[width=\textwidth]{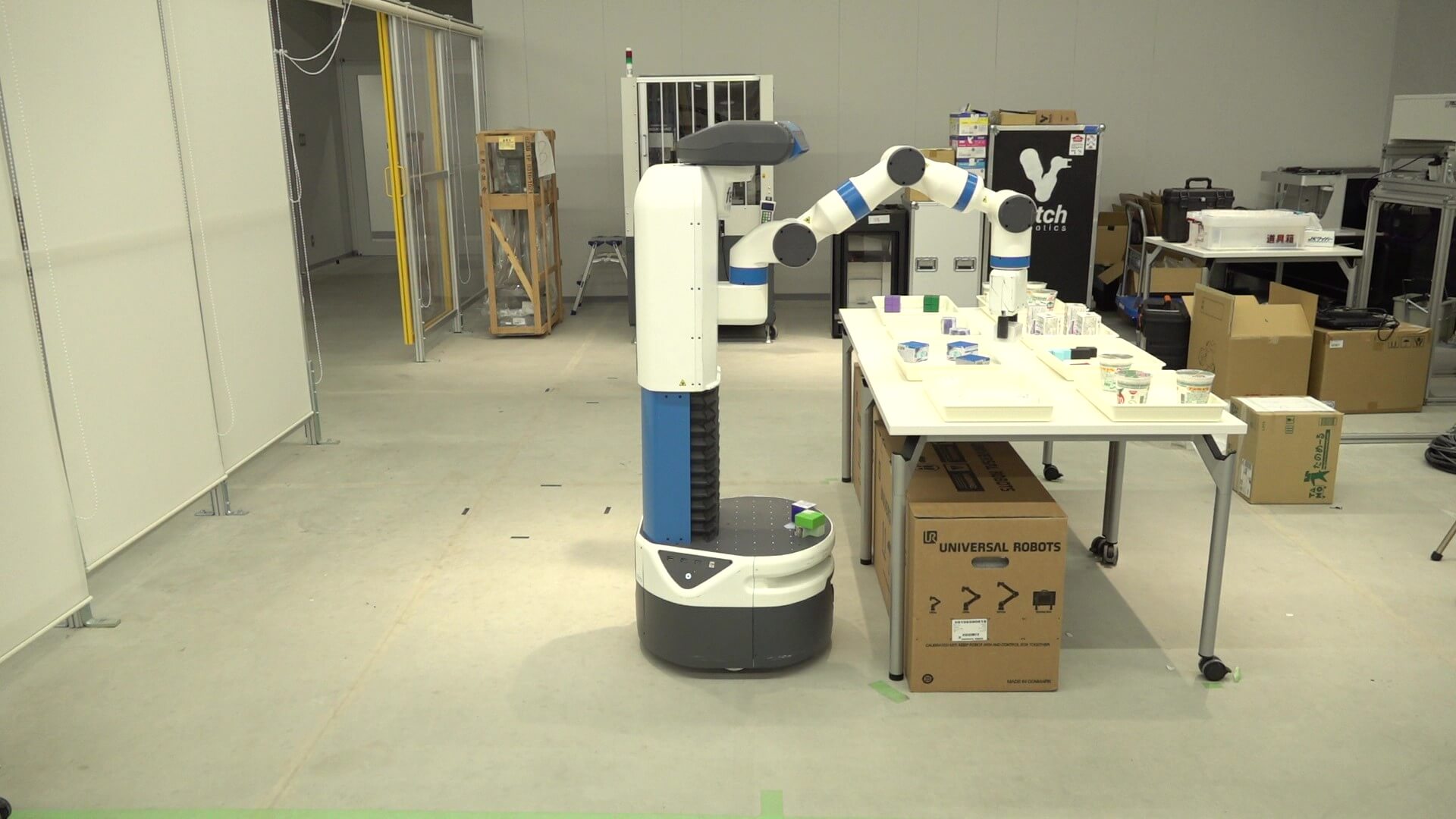}
    \end{subfigure}
    \caption{(Left and Middle) Move to $P_{B1} \cap P_{B2}$ to pick up parts from $tray_1$ and $tray_2$. (Right) Move to $P_{B4}$ to pick up a part from $tray_4$.}
    \label{fig:exper2}
\end{figure}
We performed two sets of experiment. The first experiment is for comparison, in order to collect the parts from three trays, we move the robot to the center of $P_{B1}, P_{B2}$ and $P_{B4}$, respectively. In each base position, the mobile manipulator picks up one object from the associated tray. In the second experiment, the mobile manipulator moves to the center of $P_{B1} \cap P_{B2}$ to pick up parts from $tray_1$ and $tray_2$, then moves to the center of $P_{B4}$ to pick up part from $tray_4$. The total operation time is reduced due to reduced sequence size.

\section{Conclusions}
Multiple pick-and-place tasks involved in the assembly factory are considered in this paper. Both the efficiency and the robustness with respect to base positioning uncertainty are of major importance for practical applications. The proposed IK query method is especially helpful for finding collision free IK solutions in complex environment. It is resolution complete for generating all the feasible base positions in a part-supply task. By incorporating the base positioning uncertainty into base sequence planning, the mobile manipulator is able to robustly finish the task even if the actual arrived position deviate from the planned position. Our experiment shows, following the planned base sequence, the operation time is reduced and the part-supply task is completed under real world uncertainty.

\bibliographystyle{ieeetr}

\bibliography{ref.bib}

\end{document}